\documentclass[runningheads]{llncs}

\usepackage{eccv}

\usepackage{eccvabbrv}

\usepackage{graphicx}
\usepackage{booktabs}

\usepackage{multirow}
\usepackage{wrapfig}
\usepackage{setspace}
\usepackage{floatflt}
\usepackage{float}
\usepackage{adjustbox}
\usepackage{tcolorbox}
\tcbuselibrary{theorems}
\newtcbtheorem{promptbox}{Prompt}%
{colback=gray!5,colframe=black,fonttitle=\bfseries}{prompt}

\usepackage{caption}
\usepackage[table]{xcolor}
\definecolor{onlinefull}{RGB}{235,245,255}
\definecolor{offlineweak}{RGB}{255,240,230}
\definecolor{onlineweak}{RGB}{235,255,235}
\definecolor{offlinefull}{RGB}{255,235,250}

\AtBeginDocument{
\addtolength{\abovedisplayskip}{-2ex}
\addtolength{\abovedisplayshortskip}{-2ex}
\addtolength{\belowdisplayskip}{-2ex}
\addtolength{\belowdisplayshortskip}{-2ex}
}

\usepackage[accsupp]{axessibility}  

\usepackage{hyperref}

\usepackage{orcidlink}

\begin{document}

\title{\includegraphics[scale=0.05, trim=0 2cm -1cm 0]{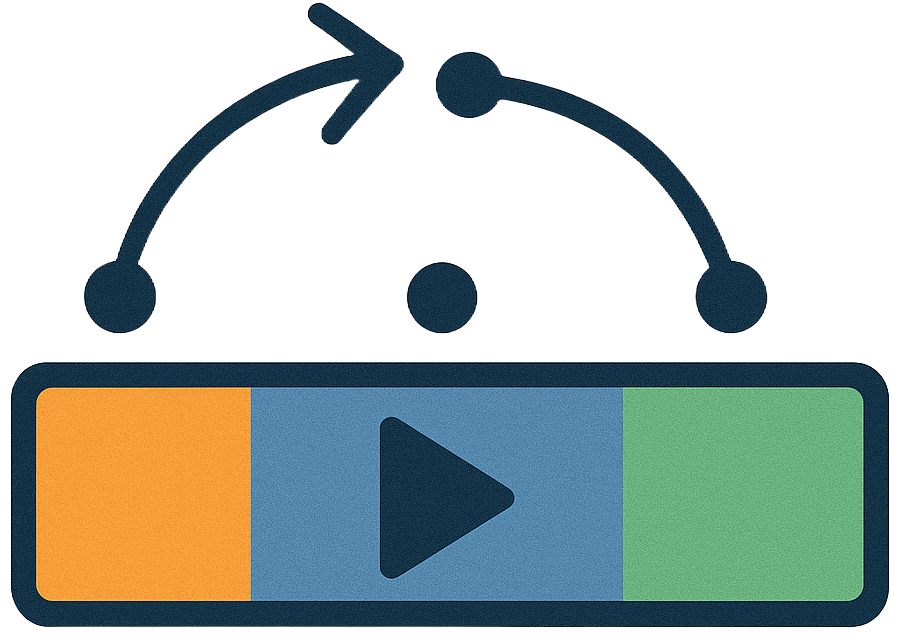} OnPoint: Offline-to-Online \\ Multi-Level Distillation for \underline{Point}-Supervised \\ \underline{On}line Temporal Action Localization} 

\titlerunning{Point-Supervised Online Temporal Action Localization}




\author{Sakib Reza\inst{1, 3}\thanks{Work primarily done during an internship at Dolby Laboratories.}\orcidlink{0000-0001-8491-0316} \and
Gauri Jagatap\inst{2}\orcidlink{0000-0001-7499-2581} \and Mohsen Moghaddam\inst{3}\orcidlink{0000-0002-3201-6010} \and \\Octavia Camps\inst{1}\orcidlink{0000-0003-1945-9172} \and Andrea Fanelli \inst{2} \orcidlink{0000-0001-9876-9050}}

\authorrunning{S. Reza et al.}




\institute{Northeastern University, Boston, MA 02115, USA \\ \email{\{reza.s, o.camps\}@northeastern.edu} \and
Dolby Laboratories, Inc., San Francisco, CA 94103, USA \\ 
\email{\{gauri.jagatap, andrea.fanelli\}@dolby.com}\\\and
Georgia Institute of Technology, Atlanta, GA 30332, USA\\
\email{\{sreza32, mohsen.moghaddam\}@gatech.edu}
}

\maketitle

\begin{abstract}
Temporal Action Localization (TAL) typically relies on segment annotations or offline access to full videos, limiting scalability and online use. We introduce Point-Supervised Online TAL (POTAL), which localizes actions in streaming videos using only one temporal point per instance. To solve POTAL, we propose OnPoint, an offline-to-online multi-level distillation framework that transfers knowledge from a point-supervised offline teacher to an online student via (i) pseudo-segment instance distillation, (ii) class-activation sequence distillation, and (iii) anticipatory window-level distillation. We further improve robustness by incorporating the original point labels into student training and by refining anchor decoding with actionness-guided attention calibration. Experiments on five datasets show OnPoint consistently outperforms strong baselines, establishing a solid foundation for POTAL. 
\renewcommand{\thefootnote}{\dag}
\footnote{Project Page: \url{https://sakibreza.github.io/OnPoint/}}
\renewcommand{\thefootnote}{\arabic{footnote}}
\end{abstract}


\section{Introduction}
\label{sec:intro}

With the rapid growth of video platforms, Temporal Action Localization (TAL)~\cite{wang2023temporal, xia2020survey}, which identifies action boundaries and class labels in untrimmed videos, has become a central task in computer vision. 
However, many real-world systems must localize actions directly from \textit{streaming video} while minimizing annotation cost. 
Examples include augmented reality (AR) task assistance for surgical training or industrial maintenance~\cite{celdran2025use, yoo2024ai}, as well as live sports analytics~\cite{shih2017survey}, surveillance~\cite{vishwakarma2013survey}, and robotics~\cite{jiang2020understanding}. 
In such settings, models must make frame-by-frame predictions without future context while operating under limited supervision, as dense action boundary annotation is often costly and infeasible at scale.

A key challenge arises from the \textit{inference regime}. 
Most TAL methods~\cite{lin2018bsn, lin2019bmn, zhang2022actionformer, xia2022dual, xia2022learning, xia2023exploring, xu2017r, wang2022rcl, gupta2025losa} assume offline access to complete videos, allowing models to leverage future context during training and inference. 
In contrast, real deployments require \textit{online inference}, where actions must be localized in streaming video as frames arrive. 
Recent Online Temporal Action Localization (OnTAL) approaches~\cite{kang2021cag, reza2024hat} address this setting but rely on full segment-level supervision during training, limiting scalability in continuously recorded environments where annotation is scarce.

\begin{wrapfigure}{t}{0.5\linewidth}
    \centering
    \vspace{-0.5cm}
    \begin{minipage}{\linewidth}
        \centering
        \includegraphics[width=0.99\linewidth]{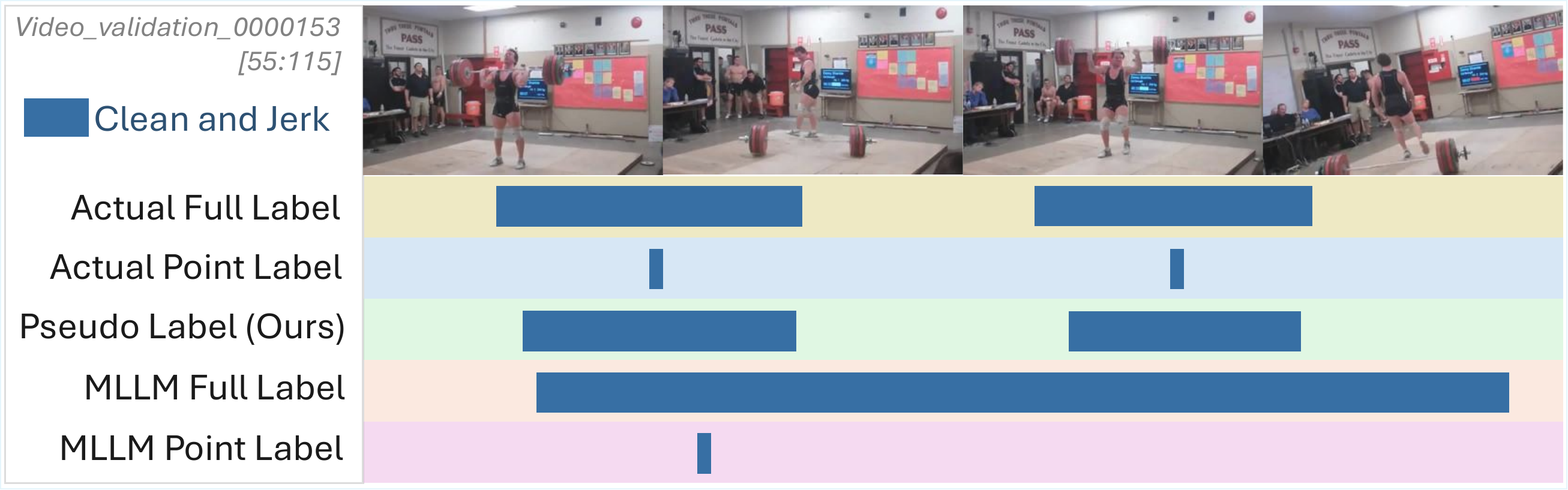}
        \vspace{-0.5cm}
        \caption{\textbf{Labeling example.} Ground truth, human point labels, pseudo segments from our point-supervised offline TAL teacher, and Gemini 2.5 Flash~\cite{comanici2025gemini} MLLM labels. Gemini often merges instances and misplaces boundaries (29.2\% avg mAP), while our teacher yields accurate pseudo labels (90.3\%) on THUMOS~\cite{idrees2017thumos}, underscoring the value of weak human annotation. (Sec. \ref{supp:annotation})}
        \label{fig:potal_llm}
    \end{minipage}

    \vspace{0.2cm}
    \begin{minipage}{\linewidth}
        \centering
        \includegraphics[width=0.99\linewidth]{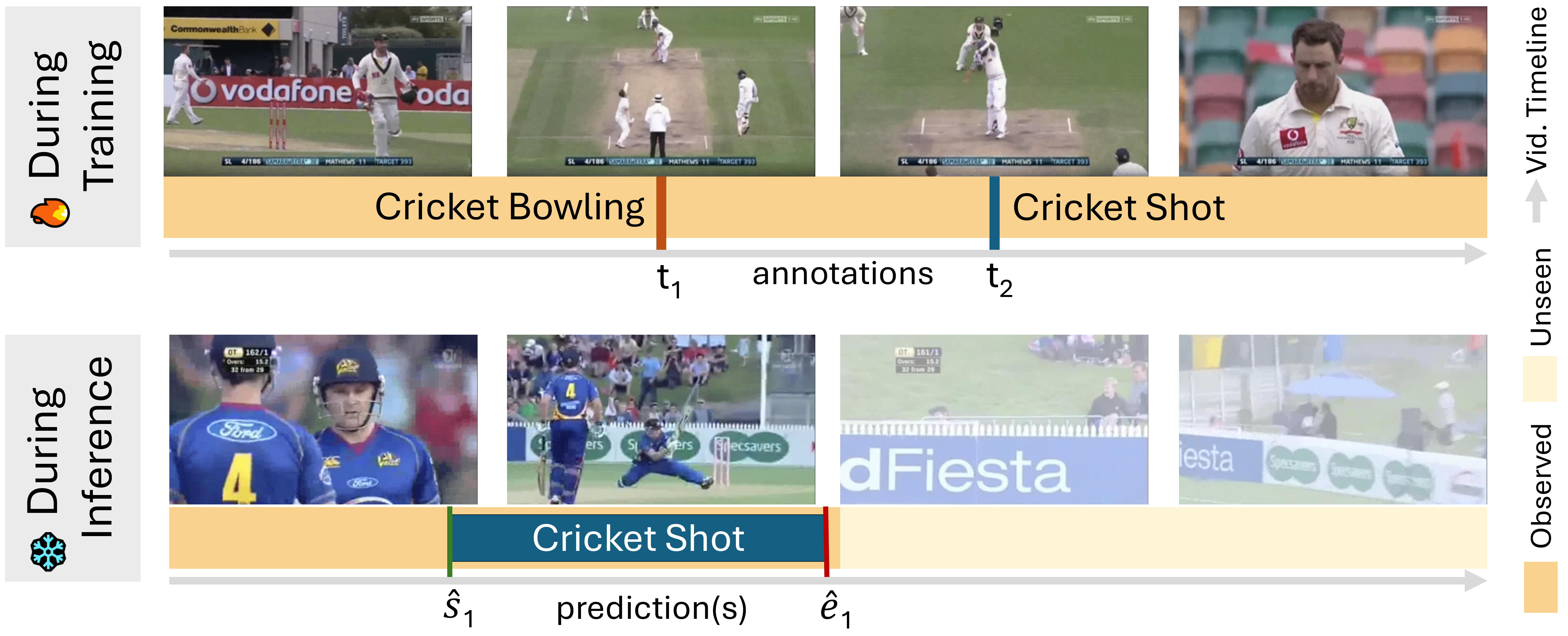}
        \vspace{-0.5cm}
        \caption{\textbf{POTAL task.} Training uses one timestamp per action instance. At test time, the model outputs action class and boundaries online, emitting each segment immediately when the action ends (no future frames).}
        \vspace{-1cm}
        \label{fig:potal_teaser}
    \end{minipage}
\end{wrapfigure}

A second challenge concerns the \textit{supervision cost}. 
Training state-of-the-art TAL or OnTAL models typically requires dense temporal annotations specifying action start and end boundaries, which are expensive to obtain. 
Recent large multimodal foundation models~\cite{nie2024slowfocus, comanici2025gemini, Reza_2025_CVPR} show promising zero-shot capabilities but remain unreliable for precise temporal action instance detection and boundary localization (Fig.~\ref{fig:potal_llm}). 
Consequently, fully automatic label generation remains impractical for many real-world deployments.

A practical alternative is \textit{point supervision}, where each action instance is annotated with a single timestamp. 
Prior work~\cite{ma2020sf} and our supplementary study (Supp. Sec. \ref{supp:annotation}) show that point labeling can reduce annotation effort by up to $6\times$ while maintaining reliability. This paradigm has been explored in Point-Supervised TAL (PS-TAL)~\cite{ma2020sf, zhang2024hr}, which learns action boundaries from sparse point labels but assumes offline access to the full video.

Despite substantial progress in PS-TAL and OnTAL, no prior work addresses their intersection. 
PS-TAL methods rely on full video context, violating the causality constraints required for streaming inference, while existing OnTAL methods require full segment supervision. 
Simply combining the two paradigms collapses supervision assumptions or breaks the online setting.

To bridge this gap, we introduce \textbf{Point-Supervised Online Temporal Action Localization (POTAL)}, a new task that combines the label efficiency of point annotations with the strict streaming constraints of online inference (Fig.~\ref{fig:potal_teaser}). 
POTAL defines a new problem setting (Sec.~\ref{sec:problem}), establishes evaluation benchmarks, and introduces both distillation-based and distillation-free baselines tailored to this scenario.

To tackle POTAL, we propose \textbf{OnPoint}, an offline-to-online distillation framework that uses a point-supervised offline TAL model as a teacher for a strictly online student. The teacher leverages full-video context to infer boundaries from sparse points and supplies structured supervision—pseudo segments, frame-wise activations, and future-window cues—enabling the student to localize actions under streaming constraints.



While offline-to-online distillation has been studied in tasks such as video instance segmentation~\cite{kim2024offline}, spatio-temporal action detection~\cite{patel2026distilling}, language modeling~\cite{liu2025offline}, and reinforcement learning~\cite{li2024guided}, it remains unexplored in temporal action localization (TAL). To the best of our knowledge, \textbf{OnPoint} is the first framework to distill knowledge from an offline TAL model into an online counterpart. It introduces a purpose-built multi-level distillation framework, with each component designed to address a specific challenge of the POTAL setting. Our key contributions are:

\begin{enumerate}

    \item \textbf{A new task setting: Point-Supervised Online TAL (POTAL).} We formulate the first temporal action localization problem that simultaneously enforces strict online inference (no future frames, no offline revision) and point-level supervision (one timestamp per instance). We provide protocols, evaluation, and strong baselines to establish POTAL as a benchmark setting.
    
    \item \textbf{OnPoint: the first offline-to-online distillation framework for TAL under point supervision.} \textbf{OnPoint} transfers knowledge from a point-supervised offline TAL model to a strictly online detector, directly addressing the two core challenges of POTAL: missing future context and sparse supervision. Training an online model using only point-level supervision is inherently difficult due to weak labels and the absence of future context. To address this, we introduce three complementary distillation signals:

\begin{enumerate}
\item \textbf{Providing structured boundary supervision through instance-level pseudo-label distillation.} Segment-level pseudo ground truth generated by the offline teacher supervises the anchor-based instance prediction module, providing structured boundary guidance (Sec.~\ref{sec:instance}).

\item \textbf{Enhancing temporal representation learning via dense frame-level distillation.} We align the student’s frame-wise predictions with the teacher’s Class Activation Sequence (CAS) via Class-Activation Sub-Sequence (CASS) supervision. While not directly used for boundary regression, this dense signal strengthens temporal representations and improves backbone feature quality (Sec.~\ref{sec:frame}).

\item \textbf{Compensating for missing future context through anticipatory window-level distillation.} To mitigate the absence of future context in online inference, we distill the teacher’s next-window action likelihoods into the student, guiding it toward imminent action boundaries and enabling timely localization without access to future frames (Sec.~\ref{sec:anticipation}).

\end{enumerate}

\item \textbf{Stabilizing online training under noisy pseudo labels via direct anchor-level point supervision.}
Because pseudo labels may be imperfect, we directly incorporate ground-truth action points as an additional anchor-level supervisory signal. This weak but reliable supervision stabilizes training and mitigates error propagation from the teacher (Sec.~\ref{sec:point}).

\item \textbf{Improving temporal proposal decoding through actionness-guided attention calibration.}
Standard anchor-based decoders rely on learned attention queries that lack explicit action priors. We introduce an actionness-guided attention calibration mechanism that leverages frame-wise actionness scores predicted by CASS to emphasize action-relevant regions and suppress background noise, improving temporal proposal accuracy (Sec.~\ref{sec:dec}).


\end{enumerate}

Extensive experiments (Sec.~\ref{sec:eval}) on five datasets show OnPoint consistently improves over strong baselines (up to $+7\%$ mAP) and is competitive with select fully supervised online methods, establishing a strong foundation for POTAL.

\section{Related Work}
\label{sec:related}

\vspace{-0.1cm}

\noindent \textbf{Temporal Action Localization (TAL)} is a fully-supervised, offline framework to  detect temporal boundaries and classify action instances. Existing one-stage approaches \cite{zhang2022actionformer, shi2022react, lin2021learning, shao2023action} jointly localize and classify actions in a single step, while two-stage pipelines \cite{qing2021temporal, lin2019bmn} first generate candidate proposals and then perform classification. Despite their impressive performance, these models rely heavily on densely annotated temporal boundaries and action labels for training. Moreover, their offline inference paradigm requires access to entire video sequences, restricting their deployment in online or streaming environments.

\noindent \textbf{Point-Supervised Temporal Action Localization (PS-TAL)} is an \emph{offline} weakly-supervised setting where each action instance is labeled by a single timestamp, providing a label-efficient alternative to full supervision and typically outperforming video-level supervision \cite{zhang2025weakly}. Existing methods fall into three lines: \textit{pseudo-label generation} that expands points to nearby frames for dense supervision \cite{ma2020sf, du2023timestamp, patsch2024tscl}; \textit{sequence/proposal modeling} that uses points to improve action completeness and calibrate snippet- or instance-level confidence \cite{lee2021learning, zhang2024hr, xia2024realigning}; and \textit{boundary refinement/global context} that searches for precise boundaries or leverages structured frameworks (e.g., optimal transport, DETR-style decoders) for long-range dependencies \cite{liu2024stepwise, liu2025boosting}. However, all PS-TAL methods assume full-video access and are therefore unsuitable for streaming scenarios that require causal, incremental predictions.

\noindent \textbf{Online Temporal Action Localization (OnTAL)}, introduced in \cite{kang2021cag}, aims to localize action boundaries and classify action instances in online settings, without relying on future frames or offline post-processing. Early approaches such as CAG-QIL \cite{kang2021cag} and SimOn \cite{tang2022simon} utilize per-frame action detection or online action detection backbones, followed by grouping consecutive predictions to generate action proposals. 2PESNet \cite{kim20222pesnet} advanced the field by introducing a two-stage framework for explicit start and end time detection at the instance level. Later, anchor-based approaches such as OAT \cite{kim2022sliding} and HAT \cite{reza2024hat} were proposed to avoid temporal fragmentation and imprecise boundaries, while ActionSwitch \cite{kang2024actionswitch} addressed the challenge of overlapping action localization. Furthermore, memory-augmented models like MATR \cite{song2024online} and HAT \cite{reza2024hat} leveraged long-term temporal dependencies to enhance instance-level reasoning. While OnTAL satisfies the online inference constraint of traditional TAL, current methods remain label-intensive, as they require full boundary annotations during training.

\section{Method}
\subsection{Problem Definition}
\label{sec:problem}

We introduce Point-Supervised Online Temporal Action Localization (POTAL), a novel task that aims to detect and classify action segments in streaming video using only point-level supervision, i.e., a single annotated timestamp $t_k$ and class label $a_k$ per action instance during training. Given a video stream $V = \{v_i\}_{i=1}^T$, the model observes only the partial sequence $V_{1:t} = \{v_i\}_{i=1}^t$ at time $t$ and must produce action proposals $(\hat{s}, \hat{e}, \hat{a}, \hat{p})$, where $\hat{s}$ and $\hat{e}$ denote predicted start and end times, $\hat{a}$ is the action class, and $\hat{p}$ is the confidence score. At inference, the model must produce proposals immediately upon detecting action ends, with no option for revision. The goal is to sequentially recover the full action set $\Psi = \{(\hat{s}_k, \hat{e}_k, \hat{a}_k)\}_{k=1}^{\hat{K}}$ in a strictly online, label-efficient setting.

\begin{figure*}[t]
\centering
\includegraphics[width=0.99\textwidth]{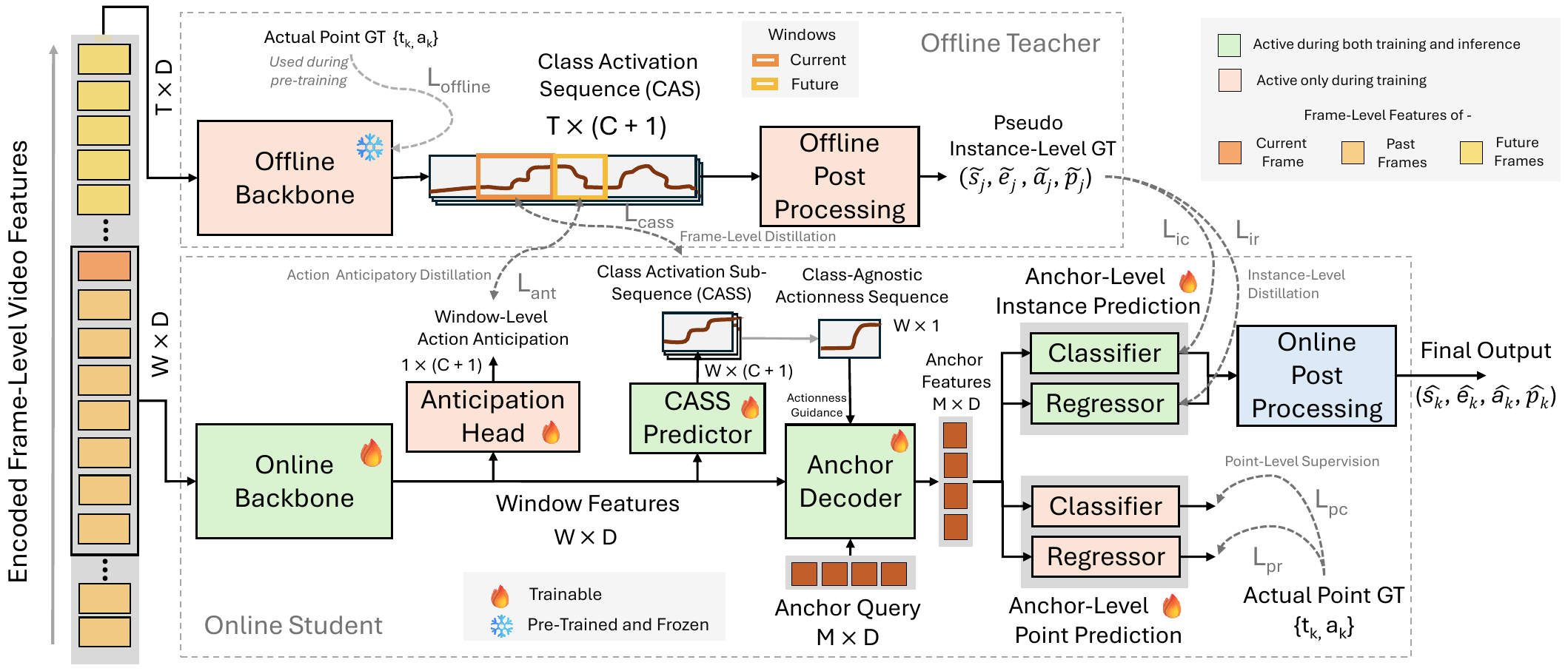} 
\caption{An overview of our proposed OnPoint framework  for POTAL task. The offline teacher model, pre-trained with point-level annotations, is frozen during training. We distill knowledge into the online student model using pseudo ground truth, frame-wise class activations, and window-level action anticipation objectives. Additionally, original point annotations are directly leveraged to supervise the online model.}
\label{fig:model}
\vspace{-0.5cm}
\end{figure*}

\subsection{Offline-to-Online Distillation Framework}

Training a standalone online model using only point-level supervision is challenging due to limited boundary cues and absence of future context. In contrast, an offline model leverages the full temporal context to infer action boundaries from sparse point annotations, generating high-quality pseudo-labels. Thus, we introduce OnPoint (Fig. \ref{fig:model}), a multi-level distillation framework comprising two core components: (1) a point-supervised offline teacher that generates pseudo-labels and class activation sequences (CAS), and (2) an online student that leverages these multi-level supervisory signals to perform online action localization using sliding-window inputs.

\noindent\textbf{Offline Teacher.} It processes the full video features $X_{\text{full}} \in \mathbb{R}^{T \times D}$, where $T$ is the number of frames and $D$ is the feature dimension, using an offline backbone to generate a class activation sequence (CAS) $A \in \mathbb{R}^{T \times (C+1)}$, with $C$ denoting the number of action classes and the extra channel representing background. Offline post-processing on $A$ yields pseudo ground truth segments $(\tilde{s}, \tilde{e}, \tilde{a}, \tilde{p})$, where $\tilde{s}$ and $\tilde{e}$ are the predicted start and end times, $\tilde{a}$ is the action label, and $\tilde{p}$ is the confidence score. The offline backbone is trained using only point-level annotations. We employ HR-Pro's backbone \cite{zhang2024hr} for sparse action datasets like THUMOS’14 \cite{idrees2017thumos}, and TSASPC's backbone \cite{du2023timestamp} for dense datasets such as EGTEA \cite{li2018eye} and HOI4D \cite{liu2022hoi4d}, given its suitability for densely annotated scenarios. We also follow the training and post-processing strategies from the mentioned prior works, with additional details provided in the supplementary material.

\noindent \textbf{Online Student.} It operates in a streaming fashion using a sliding window mechanism. At each time step $t$, it collects the current frame and the preceding $W - 1$ frames to form a feature window $X_t \in \mathbb{R}^{W \times D}$, where $W$ is the window size and $D$ is the feature dimension. This input is passed through an online backbone built with standard Transformer encoders \cite{vaswani2017attention} to capture temporal dependencies, producing intermediate representations $F_t \in \mathbb{R}^{W \times D}$.

These features are then used by three key modules: (1) a Class Activation Subsequence (CASS) predictor that estimates frame-wise class scores, (2) a window-level anticipation head that predicts potential future actions, and (3) an anchor decoder that generates anchor-level features for a set of predefined temporal anchors within the window. The anchor features are further processed by classification and regression heads, followed by online post-processing to produce final action predictions.

\noindent\textbf{Training signals.} The training of the online model is guided by the offline teacher in a multi-level distillation manner: instance-level pseudo ground truth supervises anchor-level classification and localization; the offline CAS supervises both the anticipation head and the CASS predictor; and the original point annotations are used to supervise anchor-level point prediction. More details on each module, as well as the training and inference procedures, are provided in the sections below. 
\vspace{2pt}

\subsection{Class-Activation Sub-Sequence (CASS) Distillation}
\label{sec:frame}
To provide dense frame-level supervision, we introduce CASS distillation, which learns to predict frame-wise action class scores within each sliding window. By aligning the online model’s predictions with the offline teacher’s Class Activation Sequence (CAS), CASS offers fine-grained temporal guidance that strengthens feature learning. This supervision enables the student to capture subtle action transitions more effectively, improving temporal reasoning when future context is unavailable.

Given the intermediate window-level feature representation $F_t \in \mathbb{R}^{W \times D}$, where $W$ is the window size and $D$ is the feature dimension, the CASS predictor maps each frame embedding to class probabilities using a lightweight two-layer feedforward network. Specifically, the network consists of two linear layers with a ReLU activation in between, projecting from $D$ to $D$, and then from $D$ to $C+1$, where $C$ is the number of action classes and the additional class accounts for the background. This results in frame-wise predictions $\hat{A}_t \in \mathbb{R}^{W \times (C+1)}$.

To ensure consistency with the predictions of the offline teacher, we supervise the predicted CASS using the corresponding CAS segment $A_t^{\text{teacher}} \in \mathbb{R}^{W \times (C+1)}$ via an ${L}_2$ loss, $\mathcal{L}_{\text{cass}} = \frac{1}{W} \sum_{i=1}^{W} \left\| \hat{A}_t[i] - A_t^{\text{teacher}}[i] \right\|_2^2.$ This loss encourages the online backbone to learn temporally precise, class-discriminative representations, facilitating more accurate online action localization.


\subsection{Action Anticipatory Distillation}
\label{sec:anticipation}

Window-level action anticipatory distillation strengthens the online student's temporal modeling by predicting which actions are likely to appear in the next $W'$ frames using cues distilled from the offline teacher’s CAS. This  signal guides the model toward imminent action boundaries, enabling timely localization despite having no access to future frames.

Given the intermediate window feature $F_t \in \mathbb{R}^{W \times D}$, the anticipation head begins by reducing the feature dimension to $\frac{D}{4}$ through a linear transformation. The reduced features are then flattened and passed through two fully connected layers: the first uses a ReLU activation, while the second uses a sigmoid activation to produce a class-wise probability vector $\hat{y} \in [0, 1]^{C+1}$, where $C$ is the number of action classes and the additional class accounts for background.

During training, supervision is provided by the corresponding future segment of the CAS generated by the offline teacher. A threshold of 0.5 is applied to binarize this CAS segment, resulting in a multi-hot target vector $y \in \{0, 1\}^{C+1}$ that indicates the set of actions expected to occur within the future window. The anticipation head is optimized using the binary cross-entropy loss, $\mathcal{L}_{\text{ant}} = - \sum_{c=1}^{C} \left[ y_c \log(\hat{y}_c) + (1 - y_c) \log(1 - \hat{y}_c) \right],$
where $y_c$ and $\hat{y}_c$ denote the teacher-derived target and predicted probability for action class $c$, respectively. This loss formulation enables the online model to benefit from future knowledge distilled from the offline model, improving its ability to reason about upcoming actions in an online streaming setup.


\subsection{Actionness-Calibrated Anchor Decoding}
\label{sec:dec}

Following prior works \cite{kim2022sliding, reza2024hat}, we adopt an anchor-based approach for action localization within our online model. At each sliding window, we define a fixed set of $M$ temporal anchors with predefined scales. The anchor decoder is built upon a standard transformer decoder \cite{vaswani2017attention} that learns $M$ anchor queries $\mathcal{Q} \in \mathbb{R}^{M \times D}$, and uses cross-attention to aggregate relevant information from the window feature $F_t \in \mathbb{R}^{W \times D}$, where $W$ is the window length and $D$ is the feature dimension. The output of the transformer decoder is a set of anchor features $Z \in \mathbb{R}^{M \times D}$ representing encoded action proposals across the window.

\begin{wrapfigure}[18]{!ht}{0.6\linewidth}
\vspace{-0.5cm}
\centering
\includegraphics[width=0.6\columnwidth]{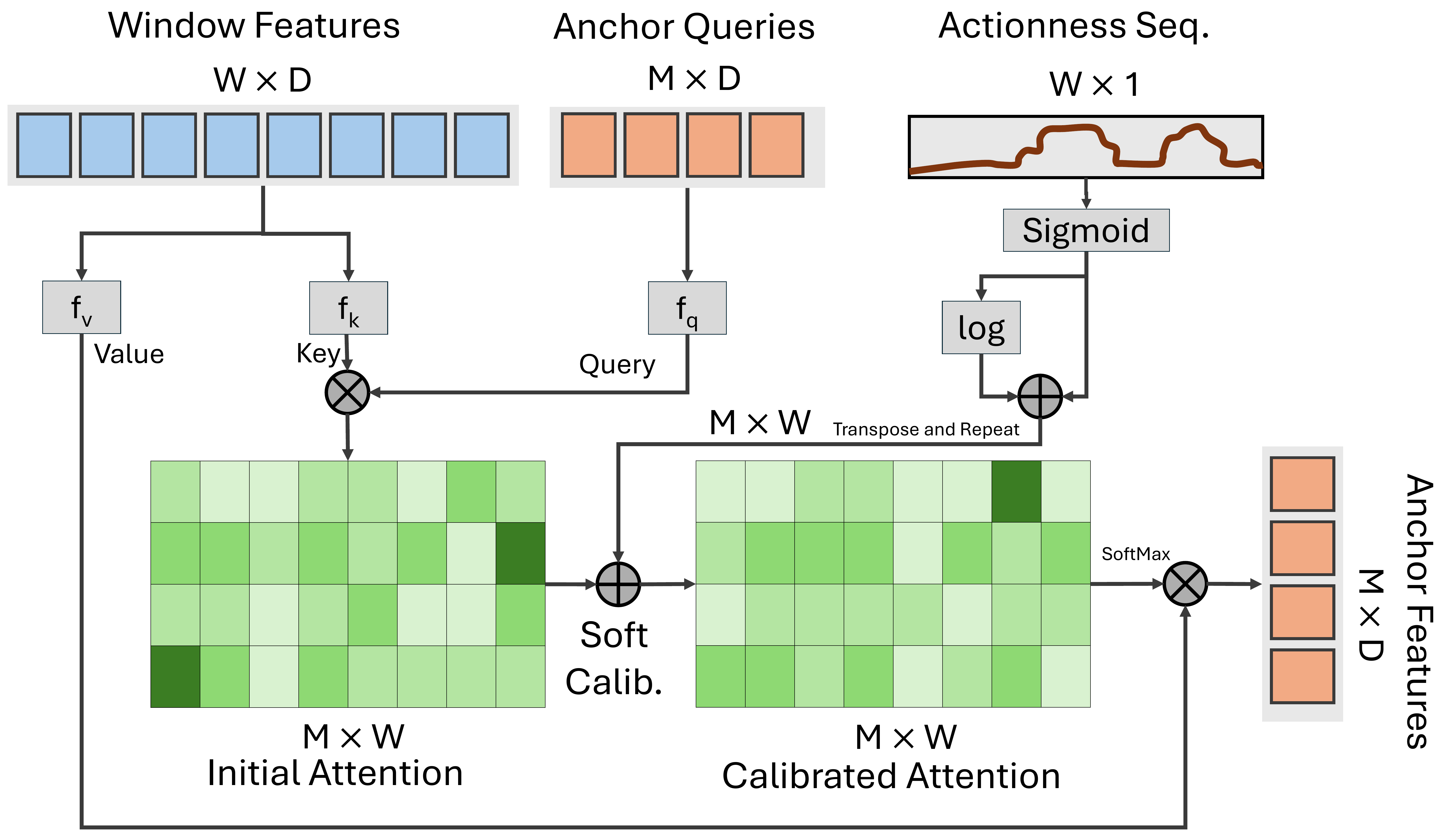} 
\caption{Actionness sequence-based attention calibration for the anchor decoder. This module introduces an intentional bias, guiding the anchor features to emphasize high-quality action information from the most relevant frames while reducing influence from less informative ones.}
\label{fig:anchor}
\end{wrapfigure}

To enhance the quality of attention in the anchor decoding process, we introduce a novel actionness-calibrated attention mechanism, which biases the attention weights using class-agnostic actionness scores derived from the Class Activation Sub-Sequence (CASS) predictor. A simple illustration of this method is presented in Figure \ref{fig:anchor} and discussed below.

Let the CASS prediction for the current window be denoted as $A_t \in \mathbb{R}^{W \times (C+1)}$, where the last channel $A_t^{(C+1)} \in \mathbb{R}^{W}$ corresponds to the background class. The actionness score sequence $r \in \mathbb{R}^{W \times 1}$ is computed as, $r = 1 - \sigma\left(A_t^{(C+1)}\right)$, where $\sigma(\cdot)$ is the sigmoid function. This formulation inversely maps low background probabilities to high actionness and vice versa.

To ensure that the calibration value is positive for high-actionness frames and negative for low-actionness frames, thus enhancing or suppressing attention accordingly, we transform the original scores using the following equation, $\bar{r} = r + \log(r)$. The resulting calibrated actionness signal $\bar{r} \in \mathbb{R}^{W \times 1}$ emphasizes frames with a higher likelihood of containing actions.

This signal is incorporated into the transformer decoder's cross-attention mechanism by adding $\bar{r}$ as a bias term to the attention logits. Specifically, if $\text{Attn}(\mathcal{Q}, \mathcal{K}, \mathcal{V})$ denotes the standard scaled dot-product attention between queries $\mathcal{Q}$, keys $\mathcal{K}$, and values $\mathcal{V}$, we modify it as, $\text{Attn}_\text{calibrated}(\mathcal{Q}, \mathcal{K}, \mathcal{V}) = \text{Softmax}\left(\frac{\mathcal{QK}^\top}{\sqrt{D}} + \bar{r}^\top\right)\mathcal{V},$ where $\mathcal{K, V} \in \mathbb{R}^{W \times D}$ are derived from the window features $F_t$. This actionness-calibrated attention mechanism emphasizes frames with strong action likelihood during anchor decoding.



\subsection{Auxiliary Anchor-Level Point Supervision}
\label{sec:point}
Our distillation framework transfers information  from the offline teacher model to the online student across multiple levels. While this supervision effectively guides the student, the pseudo ground truth generated by the teacher is inherently imperfect, allowing errors to propagate. To make the framework more robust, we directly incorporate the actual point-level ground truth into the online model’s training, ensuring stability against teacher-induced noise.

We introduce an Anchor-Level Point Prediction module that leverages point annotations as a weak yet reliable supervisory signal. The module contains a classification head to determine whether an action point lies within an anchor and its class, and a regression head to estimate the normalized distance between the anchor center and the point location. Both heads use a simple two-layer neural network with a ReLU activation in between. Based on prior evidence \cite{ma2020sf} and our pilot study (see supplementary material), point annotations collected in naturalistic settings tend to follow a Gaussian-like distribution centered around the action’s midpoint. Leveraging this prior, the model is encouraged to reward anchors centered around action points and penalize those farther away.

The total point prediction loss is defined as $\mathcal{L}_{\text{pnt}} = \mathcal{L}_{\text{pc}} + \mathcal{L}_{\text{pr}}$, where $\mathcal{L}_{\text{pc}}$ is the cross-entropy classification loss, and the corresponding regression loss is formulated as $\mathcal{L}_{\text{pr}} = \frac{1}{N_p} \sum_{j=1}^{N_p} \left| \hat{d}_j - \frac{|c_a - p_j|}{l_a} \right|,$ where $\hat{d}_j$ denotes the predicted normalized distance, $c_a$ is the anchor center, $p_j$ is the annotated point, and $l_a$ is the anchor length. This auxiliary point supervision complements the teacher’s guidance, yielding a more stable and noise-tolerant training process.

\subsection{Anchor-Level Instance Supervision}
\label{sec:instance}
In our distillation framework, the pseudo ground truth serves as direct supervision for the anchor-level instance prediction module, which is responsible for generating action proposals. Following the anchor-based approach of \cite{kim2022sliding}, this module employs two parallel heads: a classification head that determines whether an action occurs within a specific anchor and predicts its class, and a regression head that estimates temporal offsets and scaling factors to refine the anchor boundaries. The overall instance-level loss is formulated as $\mathcal{L}_{\text{ins}} = \mathcal{L}_{\text{ic}} + \mathcal{L}_{\text{ir}}$, where $\mathcal{L}_{\text{ic}}$ and $\mathcal{L}_{\text{ir}}$ represent the classification and regression losses, respectively. These objectives jointly encourage the model to identify the most relevant anchors and precisely localize their temporal extents. Throughout training, pseudo ground-truth labels produced by the offline teacher are used to supervise this instance prediction process.

\subsection{Training and Inference Strategies}

The total training objective is defined as a weighted combination of four loss components, $\mathcal{L}_{\text{total}} = \alpha \mathcal{L}_{\text{ins}} + \beta \mathcal{L}_{\text{cass}} + \gamma \mathcal{L}_{\text{ant}} + \delta \mathcal{L}_{\text{pnt}}.$ Here, $\mathcal{L}_{\text{ins}}$ is the anchor-level instance prediction loss, $\mathcal{L}_{\text{cass}}$ is the class-activation sub-sequence loss, $\mathcal{L}_{\text{ant}}$ is the anticipation loss, and $\mathcal{L}_{\text{pnt}}$ represents the point prediction loss. Definitions of these terms are provided in earlier sections. Empirically, we observe that setting $\alpha = \beta = \delta = 1$ yields the best results, and recommend tuning only $\gamma$, which governs the anticipation loss. This choice significantly reduces the hyperparameter tuning overhead. 


During inference, following \cite{kim2022sliding}, our framework generates action proposals solely from anchor-level instance predictions, denoted as $\{(\hat{s}_k, \hat{e}_k, \hat{a}_k, \hat{p}_k)\}_{k=1}^{\hat{K}}$. To reduce redundant outputs in the online setting, we apply Online Non-Maximum Suppression (ONMS) \cite{song2024online}. At each timestamp, ONMS discards proposals that have high temporal overlap with previously selected ones. Moreover, proposals with predicted end times beyond the current timestamp ($t < \hat{e}$) are also removed to ensure future, potentially more reliable predictions are not ignored.

\section{Experiments}

\subsection{Experimental Setup}

\paragraph{Datasets.} We evaluate on five benchmarks using only point-level supervision for training. (1) \textbf{THUMOS’14} \cite{idrees2017thumos}: 413 untrimmed sports videos with 20 actions, averaging $\sim$15 instances per video and large variation in action/video duration. (2) \textbf{EGTEA} \cite{li2018eye}: 28 hours of egocentric kitchen videos (86 sessions, 32 subjects) annotated with 22 fine-grained actions, averaging $\sim$180 instances and 15.6 categories per video. (3) \textbf{HOI4D-Office} \cite{reza2023enhancing, liu2022hoi4d}: 553 egocentric videos covering 12 office actions, with $\sim$15.5 instances per video, providing a structured evaluation for egocentric action understanding. (4) \textbf{FineAction} \cite{liu2022fineaction} and (5) \textbf{EPIC-Kitchens-100} \cite{damen2022rescaling}: two additional large-scale benchmarks covering more diverse and challenging action scenarios. For THUMOS’14, we use point annotations from \cite{lee2021learning} and generate point annotations for all other datasets using the same procedure.

\paragraph{Implementation Details.} We primarily report standard mAP at multiple tIoU thresholds \cite{kim2022sliding}; instance-level F1 \cite{kang2024actionswitch} is additionally reported in the supplement. For features, we follow dataset protocols: THUMOS uses two-stream TSN features (Kinetics-pretrained) as in \cite{kim2022sliding}; EGTEA uses Kinetics-pretrained I3D at 24 FPS with stride 12 \cite{reza2024hat}; and HOI4D-Office Tools uses pre-extracted CLIP-ViT features \cite{reza2023enhancing}. We adopt the offline backbone and post-processing of \cite{zhang2024hr} for THUMOS and \cite{du2023timestamp} for EGTEA/HOI4D-O. Our online model uses a $D{=}1024$ transformer encoder (3 blocks, 8 heads) and an anchor decoder (5 blocks, 4 heads). 
For sliding-window and anticipation, we use $W{=}64$, $M{=}6$, $\{4,8,16,32,48,64\}$, $W'{=}16$ for THUMOS; $W{=}24$, $M{=}7$, $\{2,4,6,8,12,16,24\}$, $W'{=}12$ for EGTEA; and $W{=}56$, $M{=}6$, $\{4,8,16,24,48,56\}$, $W'{=}16$ for HOI4D-O. We set $\gamma{=}1.0$ for THUMOS and $0.8$ otherwise. For training, we use Adam (learning rate $1{\times}10^{-4}$, weight decay $1{\times}10^{-4}$). Additional details are in the supplement.

\paragraph{Baselines.} As the first work to study POTAL, we establish both distillation-based and distillation-free baselines. For distillation-based baselines, we first use strong offline point-supervised TAL models to generate pseudo segments and then train state-of-the-art OnTAL models using these pseudo labels. We explored multiple offline teacher models (Table~\ref{tab:offback}) and found HR-Pro~\cite{zhang2024hr} effective for sparse-action scenarios such as THUMOS'14, while TSASPC~\cite{du2023timestamp} performs better for dense or procedural settings like EGTEA and HOI4D-O. As online backbones, we evaluate OAT \cite{kim2022sliding}, MATR \cite{song2024online}, and OAT-ONMS (OAT augmented with MATR’s NMS to improve proposal quality). We additionally include a distillation-free baseline trained primarily with anchor-level point supervision (details in Supp. Sec.~\ref{app:method}).

\begin{table*}[!t]
\centering
\caption{Comparison of model performance on THUMOS'14 dataset. We include results for offline fully-supervised, offline point-supervised, and online fully-supervised methods for reference. (* indicates our produced baselines.)}
\vspace{-0.3cm}
\scriptsize
\resizebox{0.9\linewidth}{!}{

\begin{tabular}{c|ccccccc|ccc}
\hline
\multirow{2}{*}{\textbf{Method}} 
& \multicolumn{7}{c|}{\textbf{mAP@tIoU (\%)}} 
& \multicolumn{3}{c}{\textbf{AVG}} \\
& 0.1 & 0.2 & 0.3 & 0.4 & 0.5 & 0.6 & 0.7 
& (0.1:0.5) & (0.3:0.7) & (0.1:0.7) \\
\hline

\rowcolor{offlinefull}
\textit{\textbf{Offline + Full}}      
&  &  &  &  &  &  &  &  &  &  \\
\rowcolor{offlinefull}
TCANet \cite{qing2021temporal}      
& - & - & 60.6 & 53.2 & 44.6 & 36.8 & 26.7 & - & 44.3 & - \\

\rowcolor{offlinefull}
AFSD \cite{lin2021learning}      
& - & - & 67.3 & 62.4 & 55.5 & 43.7 & 31.1 & - & 52.0 & - \\

\rowcolor{offlinefull}
React \cite{shi2022react} 
& - & - & 69.2 & 65.0 & 57.1 & 47.8 & 35.6 & - & 55.0 & - \\

\rowcolor{offlinefull}
ASL \cite{shao2023action} 
& - & - & 83.1 & 79.0 & 71.7 & 59.7 & 45.8 & - & 67.9 & - \\

\hline

\rowcolor{offlineweak}
\textit{\textbf{Offline + Point}}      
&  &  &  &  &  &  &  &  &  &  \\
\rowcolor{offlineweak}
LACP \cite{lee2021learning}        
& 75.7 & 71.4 & 64.6 & 56.5 & 45.3 & 34.4 & 21.8 & 62.7 & 44.5 & 52.8 \\

\rowcolor{offlineweak}
SMBD \cite{liu2024stepwise}  
& - & - & 66.0 & 57.9 & 47.0 & 36.0 & 22.0 & 64.2 & 45.7 & - \\

\rowcolor{offlineweak}
RCTSI \cite{xia2024realigning} 
& 82.3 & 77.6 & 70.1 & 60.0 & 49.4 & 37.6 & 24.5 & 67.9 & 48.3 & 57.4 \\

\rowcolor{offlineweak}
TSASPC \cite{du2023timestamp} 
& 70.2 & 68.8 & 67.1 & 63.2 & 58.0 & 51.6 & 43.1 & 65.5 & 56.6 & 60.4 \\

\rowcolor{offlineweak}
HR-Pro \cite{zhang2024hr} 
& 85.6 & 81.6 & 74.3 & 64.3 & 52.2 & 39.8 & 24.8 & 71.6 & 51.1 & 60.3 \\

\rowcolor{offlineweak}
QROT \cite{liu2025boosting}   
& - & - & 73.1 & 64.4 & 54.3 & 41.3 & 27.4 & 72.3 & 52.1 & - \\

\hline

\rowcolor{onlinefull}
\textit{\textbf{Online + Full}}      
&  &  &  &  &  &  &  &  &  &  \\
\rowcolor{onlinefull}
CAG-QIL \cite{kang2021cag}   
& - & - & 44.7 & 37.6 & 29.8 & 21.9 & 14.5 & - & 29.7 & - \\

\rowcolor{onlinefull}
2PESNET \cite{kim20222pesnet}      
& - & - & 47.4 & 39.8 & 31.4 & 21.8 & 14.0 & - & 30.9 & - \\

\rowcolor{onlinefull}
SimOn \cite{tang2022simon}        
& - & - & 57.0 & 47.5 & 37.3 & 26.6 & 16.0 & - & 36.9 & - \\

\rowcolor{onlinefull}
OAT-OSN \cite{kim2022sliding}     
& - & - & 63.0 & 56.7 & 47.1 & 36.3 & 20.0 & - & 44.6 & - \\

\rowcolor{onlinefull}
HAT \cite{reza2024hat}     
& - & - & 62.0 & 57.0 & 48.0 & 36.5 & 20.7 & - & 44.8 & - \\

\rowcolor{onlinefull}
MATR-ONMS \cite{song2024online}       
& - & - & 70.3 & 62.7 & 52.1 & 38.6 & 23.7 & - & 49.5 & - \\

\rowcolor{onlinefull}
OAT-ONMS \cite{kim2022sliding, song2024online}*       
& - & - & 66.1 & 61.3 & 52.4 & 41.6 & 26.0 & - & 49.5 & - \\

\hline

\rowcolor{onlineweak}
\textit{\textbf{Online + Point}}      
&  &  &  &  &  &  &  &  &  &  \\
\rowcolor{onlineweak}
Distillation-Free Baseline*       
& 59.9 & 45.8 & 31.8 & 18.4 & 10.3 & 5.6 & 2.0 & 33.3 & 13.6 & 24.8 \\

\rowcolor{onlineweak}
HR-Pro + OAT-OSN *       
& 59.1 & 55.8 & 49.3 & 41.5 & 29.7 & 16.3 & 7.1 & 47.1 & 28.8 & 37.0 \\

\rowcolor{onlineweak}
HR-Pro + HAT*       
& 58.4 & 55.2 & 48.9 & 39.3 & 28.2 & 16.5 & 6.6 & 46.0 & 27.9 & 36.2 \\

\rowcolor{onlineweak}
HR-Pro + MATR-ONMS*      
& 66.8 & 62.9 & 56.5 & 48.2 & 36.7 & 25.5 & 14.6 & 54.2 & 36.3 & 44.5 \\

\rowcolor{onlineweak}
HR-Pro + OAT-ONMS*      
& 66.6 & 63.7 & 58.7 & 49.3 & 40.1 & 28.7 & 16.6 & 55.7 & 38.7 & 46.3 \\

\rowcolor{onlineweak}
\textbf{OnPoint (Ours)}      
& \textbf{73.6} & \textbf{70.3} & \textbf{63.9} & \textbf{56.3} & \textbf{45.2} 
& \textbf{30.8} & \textbf{17.5} 
& \textbf{61.9} & \textbf{42.7} & \textbf{51.1} \\

\hline
\end{tabular}
}
\vspace{-0.2cm}
\label{tab:thumos}
\end{table*}

\begin{table*}[!t]
\scriptsize
\centering
\caption{Comparisons of performance on EGTEA and HOI4D-O datasets. (* indicates our produced baselines.)}
\vspace{-0.3cm}
\resizebox{0.99\linewidth}{!}
{
\begin{tabular}{l|cccccc|cccccc}
\hline
\multirow{3}{*}{\textbf{Method}} 
& \multicolumn{6}{c|}{\textbf{EGTEA}} 
& \multicolumn{6}{c}{\textbf{HOI4D-O}} \\
& \multicolumn{6}{c|}{mAP@tIoU (\%)} 
& \multicolumn{6}{c}{mAP@tIoU (\%)} \\
& 0.1 & 0.2 & 0.3 & 0.4 & 0.5 & AVG[0.1:0.5] 
& 0.1 & 0.2 & 0.3 & 0.4 & 0.5 & AVG[0.1:0.5] \\
\hline

\rowcolor{offlineweak}
\textit{\textbf{Offline + Point}} 
&  &  &  &  &  & 
&  &  &  &  &  &  \\
\rowcolor{offlineweak}
HR-Pro \cite{zhang2024hr} 
& 18.5 & 15.3 & 12.3 & 9.3 & 6.9 & 12.5
& 69.4 & 64.6 & 57.4 & 49.7 & 41.3 & 56.5 \\
\rowcolor{offlineweak}
TSASPC \cite{du2023timestamp} 
& 48.6 & 46.0 & 41.5 & 36.4 & 30.6 & 40.6 
& 77.7 & 73.5 & 68.0 & 59.8 & 51.2 & 66.0 \\

\hline

\rowcolor{onlinefull}
\textit{\textbf{Online + Full}} 
&  &  &  &  &  & 
&  &  &  &  &  &  \\
\rowcolor{onlinefull}
OAT-OSN \cite{kim2022sliding} 
& 24.2 & 22.7 & 20.5 & 16.3 & 10.9 & 18.9 
& 55.5 & 53.2 & 48.9 & 42.0 & 33.8 & 46.7 \\

\rowcolor{onlinefull}
HAT \cite{reza2024hat} 
& 27.3 & 25.3 & 21.6 & 16.9 & 12.8 & 20.8 
& 59.3 & 55.9 & 49.0 & 42.1 & 32.9 & 47.8 \\

\rowcolor{onlinefull}
MATR-ONMS \cite{song2024online} 
& 21.5 & 19.5 & 15.9 & 12.6 & 8.6 & 15.6 
& 57.9 & 54.5 & 48.8 & 42.9 & 34.6 & 47.7 \\

\rowcolor{onlinefull}
OAT-ONMS \cite{kim2022sliding, song2024online}* 
& 30.1 & 28.5 & 25.4 & 20.0 & 14.2 & 23.7 
& 57.2 & 54.7 & 49.9 & 44.5 & 37.7 & 48.8 \\

\hline

\rowcolor{onlineweak}
\textit{\textbf{Online + Point}} 
&  &  &  &  &  & 
&  &  &  &  &  &  \\
\rowcolor{onlineweak}
Distillation-Free Baseline* 
& 26.0 & 19.9 & 12.9 & 7.8 & 4.8 & 14.3 
& 45.3 & 28.9 & 16.7 & 7.9 & 4.0 & 20.6 \\

\rowcolor{onlineweak}
TSASPC + OAT-OSN* 
& 24.4 & 21.5 & 17.5 & 12.0 & 7.1 & 16.5 
& 55.6 & 49.8 & 42.0 & 33.3 & 24.2 & 41.0 \\

\rowcolor{onlineweak}
TSASPC + HAT* 
& 23.7 & 21.4 & 17.5 & 12.5 & 8.0 & 16.6 
& 55.6 & 50.4 & 43.0 & 36.1 & 24.7 & 42.0 \\

\rowcolor{onlineweak}
TSASPC + MATR* 
& 20.6 & 18.4 & 15.1 & 11.1 & 7.1 & 14.5 
& 55.8 & 51.5 & 44.8 & 32.9 & 22.7 & 41.6 \\

\rowcolor{onlineweak}
TSASPC + OAT-ONMS* 
& 27.0 & 24.4 & 20.8 & 16.0 & 10.3 & 19.7 
& 55.9 & 52.6 & 45.2 & 35.2 & 22.6 & 42.3 \\

\rowcolor{onlineweak}
\textbf{OnPoint (Ours)} 
& \textbf{30.9} & \textbf{28.6} & \textbf{24.5} & \textbf{18.9} & \textbf{12.6} & \textbf{23.1} 
& \textbf{58.2} & \textbf{53.5} & \textbf{47.4} & \textbf{37.4} & \textbf{26.4} & \textbf{44.6} \\

\hline
\end{tabular}
}
\label{tab:egteahoi4d}
\vspace{-0.1cm}
\end{table*}

\subsection{Evaluation on POTAL}
\label{sec:eval}

We assess our approach on the POTAL task by comparing it with our baselines and existing fully-supervised, point-supervised, and online TAL methods.

\noindent \textbf{THUMOS'14 and FineAction.} As shown in Table \ref{tab:thumos}, our method outperforms all baselines on the THUMOS'14 test set, achieving an average mAP gain of 4.8\% across tIoU thresholds of 0.1--0.7, with improvements reaching up to 7.0\% at individual thresholds. It also surpasses several early fully-supervised Online TAL methods and performs competitively with some fully-supervised offline and point-supervised TAL approaches. On the more challenging FineAction benchmark (Supplementary Section \ref{sup_sec:sota}), featuring fine-grained, dense, and overlapping actions, OnPoint achieves 7.4\% average mAP, outperforming the best baseline by 2.1\%.

\noindent \textbf{EGTEA, HOI4D-O, and EPIC-Kitchens-100.} To validate generalizability across egocentric and procedural domains, we evaluate on EGTEA, HOI4D-O, and EPIC-Kitchens-100. As shown in Table \ref{tab:egteahoi4d}, our method achieves consistent average mAP gains of 3.4\% on EGTEA and 2.3\% on HOI4D-O over tIoU thresholds of 0.1--0.5. Similar gains are observed on the larger EPIC-Kitchens-100 benchmark (Supplementary Section \ref{sup_sec:sota}), where the average mAP increases from 8.5\% to 10.5\%.

\subsection{Model Analysis}

\noindent \textbf{Impact of Multi-Level Distillation.}
Our multi-level offline-to-online distillation framework includes frame-level class activation sub-sequence (CASS) alignment and window-level anticipation in addition to instance-level pseudo supervision. Removing these two components leads to a 4.4\% average performance drop (Table \ref{main:abl1}). Ablating either individually also degrades performance, confirming the value of each. Notably, removing CASS also disables the actionness-calibrated attention module, which relies on CASS inputs.

\begin{wraptable}[23]{r}{0.56\linewidth}
\centering
\scriptsize
\vspace{-1cm}
\caption{Impact of proposed Window-Level Anticipation Distillation (WAD), Actionness Sequence-based Attention Calibration (ASAC), Class-Activation Sub-Sequence Distillation (CASD) on THUMOS'14 dataset.}
\begin{adjustbox}{width=\linewidth}
\begin{tabular}{c|ccccc|c}
\hline
\multirow{2}{*}{Method} & \multicolumn{5}{c|}{mAP@tIoU (\%)} & AVG \\
& 0.1 & 0.2 & 0.3 & 0.4 & 0.5 & [0.1:0.5] \\
\hline
\textbf{Proposed} & \textbf{73.6} & \textbf{70.3} & \textbf{63.9} & \textbf{56.3} & \textbf{45.2} & \textbf{61.9} \\
w/o WAD & 71.1 & 67.9 & 62.3 & 54.6 & 44.3 & 60.0\\
w/o ASAC & 72.6 & 69.2 & 63.1 & 53.6 & 41.3 & 59.9\\
w/o CASD \& ASAC & 71.1 & 67.7 & 61.7 & 52.3 & 40.0 & 58.5\\
w/o ASAC + WAD & 69.2 & 65.3 & 60.0 & 52.8 & 42.5 & 58.0\\
w/o CASD \& ASAC + WAD & 69.2 & 65.1 & 60.2 & 52.1 & 40.8 & 57.5\\
\hline
\end{tabular}
\end{adjustbox}
\label{main:abl1}
\caption{Impact of Anchor-level point supervision on THUMOS'14 dataset. APC = Anchor-Level Point Classification Head, APR = Anchor-Level Point Regression Head (* indicates that the corresponding loss is used as the primary loss (distillation-free setup) rather than an auxiliary loss)}
\begin{adjustbox}{width=\linewidth}
\begin{tabular}{c|ccccc|c}
\hline
\multirow{2}{*}{Method} & \multicolumn{5}{c|}{mAP@tIoU (\%)} & AVG \\
& 0.1 & 0.2 & 0.3 & 0.4 & 0.5 & [0.1:0.5] \\
\hline
\textbf{Proposed} & \textbf{73.6} & \textbf{70.3} & \textbf{63.9} & \textbf{56.3} & \textbf{45.2} & \textbf{61.9} \\
\hline
w/o APC & 71.3 & 67.6 & 61.0 & 53.2 & 41.1 & 58.8 \\
w/o APR & 72.9 & 69.0 & 62.4 & 52.8 & 40.3 & 59.5\\
w/o APC + APR & 70.1 & 67.3 & 61.1 & 52.5 & 40.3 & 58.4 \\
\hline
only APC + APR* & 59.9 & 45.8 & 31.8 & 18.4 & 10.3 & 33.3 \\
only APC* & 58.7 & 46.2 & 31.0 & 15.4 & 6.0 & 31.5 \\
\hline
\end{tabular}
\end{adjustbox}
\label{main:abl2}
\hfill


\end{wraptable}


\noindent \textbf{Impact of Actionness Calibrated Attention.}
Removing the actionness sequence-based attention calibration (ASAC) from the anchor decoder reduces performance by 2.0\% (Table~\ref{main:abl1}). When removed from different combinations within our framework, performance consistently drops, highlighting its importance in refining anchor features through improved attention calibration. We further evaluate several variants of ASAC in Table~\ref{main:asac}. The main formulation described in Sec.~\ref{sec:dec} ($\bar{r} = r + \log(r)$) achieves the best performance, while ablation variants, including suppression-only ($\bar{r} = \log r$), enhancement-only ($\bar{r} = r$), simple scaling (multiplying $r$ directly with the attention map), and direct suppression ($\bar{r} = -\sigma(A_t^{(C+1)})$), all lead to lower performance, highlighting the effectiveness of our proposed attention calibration formulation.

\begin{wraptable}[14]{r}{0.56\linewidth}
\centering
\begin{adjustbox}{width=\linewidth}
\centering
\includegraphics[width=\columnwidth]{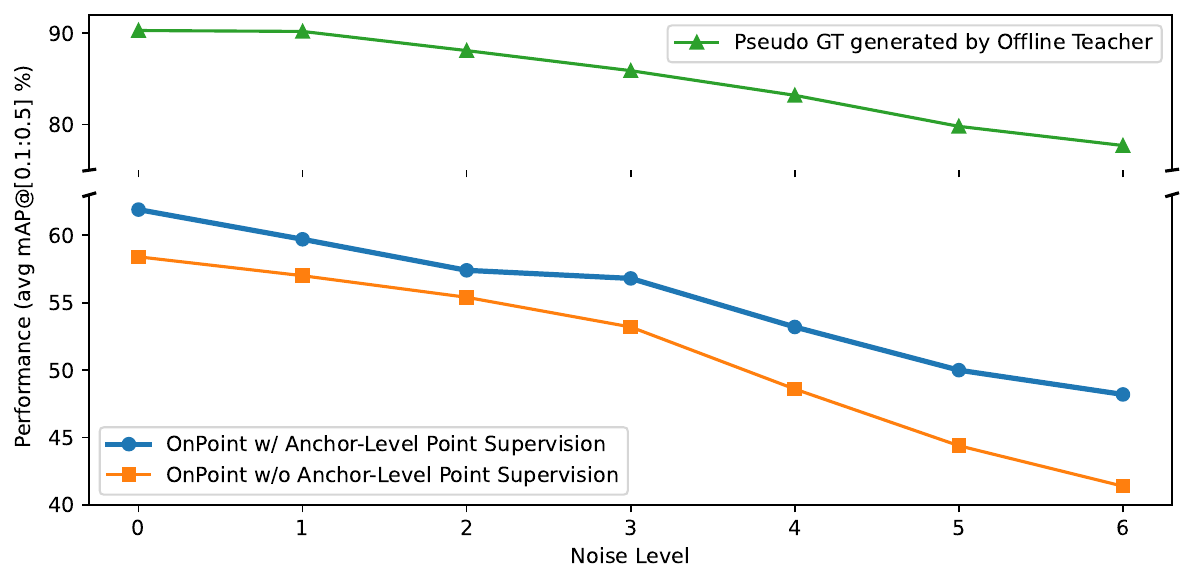} 
\end{adjustbox}
\vspace{-0.8cm}
\captionof{figure}{Effect of offline-teacher noise on the online student. Anchor-level point supervision improves robustness to  added increasing label noise and yields more stable performance than training without it by combining teacher guidance with point-level ground truth.}

\label{fig:noise}
\vspace{-0.3cm}
\end{wraptable}

\noindent \textbf{Impact of Anchor-Level Point Prediction.}
Although auxiliary, the anchor-level point prediction improves training, with a 3.5\% performance drop when removed (Table \ref{main:abl2}). Excluding either its classification or regression branch also leads to noticeable degradation, underscoring its contribution.

Moreover, this point-level supervision enhances the model’s robustness against error propagation from the offline teacher. As illustrated in Figure \ref{fig:noise}, we introduced uniform noise at varying levels into the class activation sequence (CAS) outputs of the offline backbone, which in turn distorted the pseudo ground truth and propagated errors to the online student through distillation. Our model, equipped with anchor-level point supervision, maintained higher stability under increasing noise and consistently outperformed its variant without this component. This robustness stems from the model’s dual reliance on both the teacher’s guidance and the original point-level annotations, allowing it to learn more reliably even when teacher signals become noisy.

\begin{figure}[t]
\centering
\includegraphics[width=1\textwidth]{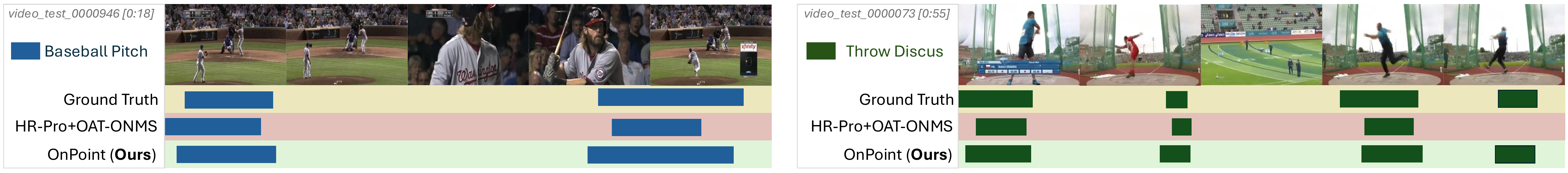} 
\caption{Qualitative comparison of action localization results on two examples from the THUMOS'14 dataset: Baseball Pitch (left) and Throw Discus (right). We compare \textbf{OnPoint} and the baseline HR-Pro+OAT-ONMS and the ground truth annotations. For Baseball Pitch, both methods detect the action; however, our approach provides more complete and accurate temporal boundaries. For Throw Discus, the baseline fails to localize one subtle instance, whereas our method successfully detects all action instances with satisfactory boundary alignment.}
\label{fig:qual1}
\vspace{-0.6cm}
\end{figure}

\begin{wraptable}[12]{r}{0.55\linewidth}
\centering
\scriptsize
\vspace{-0.5cm}



\begin{minipage}{\linewidth}
\centering
\vspace{-0.8cm}
\captionof{table}{Performance comparison of our framework using different offline backbone models, evaluated with and without our proposed components on THUMOS'14.}
\begin{adjustbox}{width=\linewidth}
\begin{tabular}{c|ccccc|c}
\hline
\multirow{2}{*}{Method} & \multicolumn{5}{c|}{mAP@tIoU (\%)} & AVG \\
& 0.1 & 0.2 & 0.3 & 0.4 & 0.5 & [0.1:0.5] \\
\hline
HR-Pro \cite{zhang2024hr} (Whole) & 66.6 & 63.7 & 58.7 & 49.3 & 40.1 & 55.7 \\
+ \textbf{Our components} & \textbf{71.5} & \textbf{68.4} & \textbf{63.0} & \textbf{53.9} & \textbf{44.3} & \textbf{60.2} \\
\hline
HR-Pro \cite{zhang2024hr} (RP) & 69.8 & 66.2 & 60.3 & 51.6 & 41.7 & 57.9 \\
+ \textbf{Our components} & \textbf{73.6} & \textbf{70.3} & \textbf{63.9} & \textbf{56.3} & \textbf{45.2} & \textbf{61.9} \\
\hline
SMBD \cite{liu2024stepwise} & 68.6 & 63.4 & 56.8 & 48.3 & 35.9 & 54.6\\
+ \textbf{Our components} & \textbf{71.3} & \textbf{66.8} & \textbf{61.3} & \textbf{50.1} & \textbf{36.6} & \textbf{57.2} \\
\hline
LACP \cite{lee2021learning} & 59.7 & 55.9 & 49.6 & 41.0 & 32.2 & 47.7\\
+ \textbf{Our components} & \textbf{64.0} & \textbf{60.3} & \textbf{54.4} & \textbf{44.4} & \textbf{32.9} & \textbf{51.2} \\
\hline
TSASPC \cite{du2023timestamp} & 61.5 & 59.9 & 55.5 & 48.8 & 38.7 & 52.9\\
+ \textbf{Our components} & \textbf{66.7} & \textbf{63.9} & \textbf{59.1} & \textbf{52.0} & \textbf{43.1} & \textbf{57.0} \\
\hline
\end{tabular}
\end{adjustbox}
\label{tab:offback}
\end{minipage}

\end{wraptable}

\noindent \textbf{Framework Generalizability Across Offline Backbones.} Table \ref{tab:offback} illustrates the generalizability of our framework when applied to a range of existing offline backbone models. Although these backbones are not part of our main contribution, our framework treats them as modular and interchangeable components, enabling straightforward integration with future, more advanced offline architectures. Across all cases, incorporating our proposed modules consistently yields superior performance compared to the same backbones combined with a simple distillation baseline. This demonstrates the robustness and adaptability of our framework, showing that it can reliably enhance diverse off-the-shelf offline models regardless of their specific architecture.


\begin{wraptable}[18]{r}{0.4\linewidth}
\centering
\scriptsize
\vspace{-1cm}

\begin{minipage}{\linewidth}
\centering
\vspace{-0.3cm}
\captionof{table}{Comparison of Actionness Sequence-based Attention Calibration (ASAC) variants on THUMOS.}
\begin{adjustbox}{width=\linewidth}
\begin{tabular}{l | c}
\hline
\multirow{2}{*}{Method} & avg mAP (\%)\\
& @tIoU [0.1:0.5]\\
\hline
\textbf{w/ ASAC (ours)} & \textbf{61.9} \\
w/o ASAC & 59.9 \\
w/ ASAC suppress only & 61.5 \\
w/ ASAC enhance only & 61.3 \\
w/ ASAC simple scaling & 58.3 \\
w/ ASAC direct suppress & 59.8 \\
\hline
\end{tabular}
\end{adjustbox}
\label{main:asac}
\end{minipage}

\vspace{0.3cm}

\begin{minipage}{\linewidth}
\centering
\includegraphics[width=\linewidth]{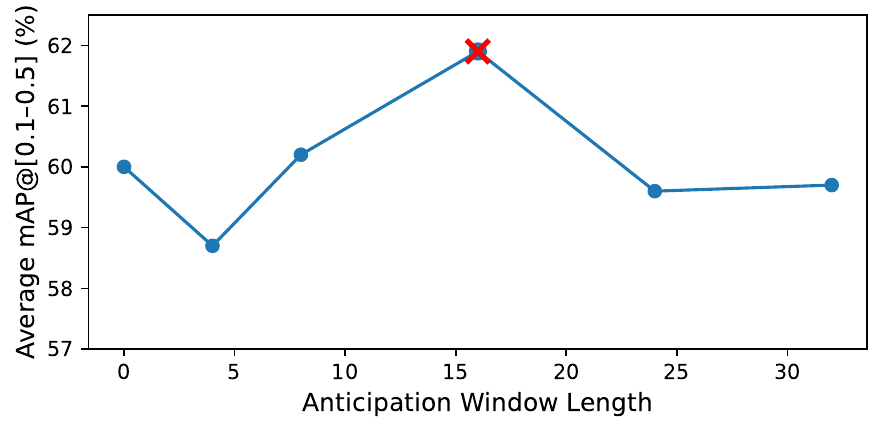}
\vspace{-0.5cm}
\captionof{figure}{Effect of anticipation window length on THUMOS14.}
\label{main:anti}
\end{minipage}

\end{wraptable}

\noindent \textbf{Additional Analysis.}
For the \textbf{anticipation distillation}, the window-length analysis in Fig.~\ref{main:anti} shows that a 16-frame anticipation window yields the best performance on THUMOS. We further compare alternative distillation strategies in the supplementary material Sec. \ref{supp:anti}, where this simple fixed-window approach performs best.
Regarding \textbf{post-processing}, our analysis (Supp. Sec.~\ref{supp:post}) shows that replacing ONMS~\cite{song2024online} with OSN~\cite{kim2022sliding} substantially reduces average mAP@tIoU[0.1:0.5] from 61.9 to 51.6, although \textbf{temporal promptness} improves, with Average Early Detection Time \cite{kim2022sliding} (AEDT)@tIoU[0.1:0.5] decreasing from $-0.17$ to $-1.53$. This highlights an inherent trade-off: ONMS achieves better localization accuracy with slightly reduced promptness, whereas OSN improves temporal responsiveness at the cost of localization performance. 
The \textbf{inference efficiency analysis} in Supp. Sec.~\ref{supp:inf} further shows that our online model remains computationally efficient (93M parameters, 2.88 GFLOPs, 312 FPS), achieving a favorable balance between speed and model capacity compared to heavier alternatives such as HAT and MATR, while remaining competitive with OAT across all efficiency metrics. 
Additional analyses, including \textbf{hyperparameter choices}, \textbf{sensitivity studies}, \textbf{CASS loss comparison}, and \textbf{instance-level F1 evaluation}, are provided in Supp. Sec.~\ref{supp:exp}.


\begin{figure}[!t]
\centerline{\includegraphics[width=0.98\linewidth]{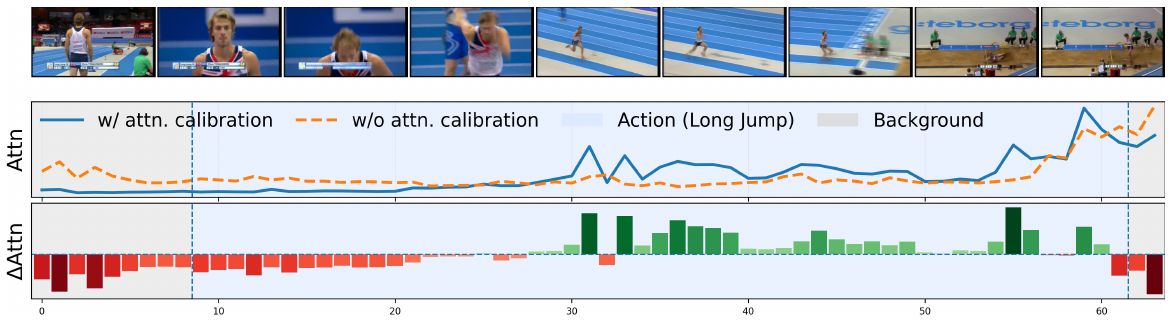}}
\caption{\textbf{Qualitative effect of attention calibration} (last decoder layer attention example on THUMOS'14). Compared to the baseline, attention calibration sharpens attention over action-relevant past frames while suppressing background, yielding more discriminative temporal context and improved downstream performance (see Supp. Sec. \ref{supp:calib} for additional analysis).}\label{fig:attention_calib}
\vspace{-2mm}
\end{figure}

\subsection{Qualitative Evaluation}

\begin{wrapfigure}[13]{ht}{0.7\textwidth}
\vspace{-0.75cm}
\centering
\includegraphics[width=0.7\textwidth]{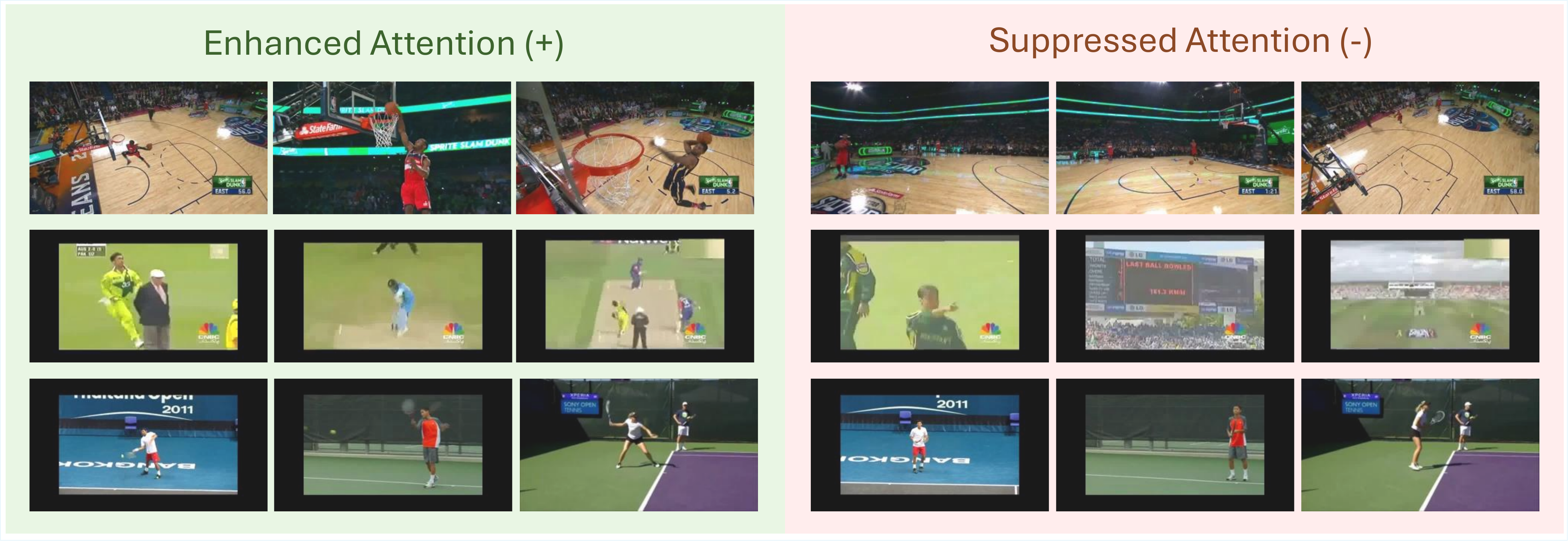} 
\caption{OnPoint's actionness-based attention calibration enhances attention in frames with key action information while suppressing attention in irrelevant segments. Examples illustrate improved attention on action-relevant regions and reduced attention in non-informative intervals.
}
\label{fig:qual2}
\vspace{-0.5cm}
\end{wrapfigure}
Figure \ref{fig:qual1} compares OnPoint against the strongest baseline, HR-Pro+OAT-ONMS, on the THUMOS'14 test set. OnPoint provides more accurate localization of action instances. In the Baseball Pitch, both methods detect the action, but ours yields tighter and more complete temporal boundaries. In the Throw Discus case, the baseline misses a subtle instance, whereas our method correctly captures all occurrences with well-aligned boundaries. Figures \ref{fig:attention_calib} and \ref{fig:qual2} qualitatively illustrate the effect of our actionness-based attention calibration, which emphasizes key action frames while suppressing irrelevant ones, producing more discriminative anchor features and improving downstream performance.
{
\parfillskip=0pt
\parskip=0pt
\par}


\section{Conclusion}

This paper introduces \textbf{Point-Supervised Online Temporal Action Localization (POTAL)}, a novel and practical problem setting for action localization in streaming videos trained with only point-level annotations. We propose \textbf{OnPoint}, which transfers full-video knowledge from an offline point-supervised teacher to a strictly online student via multi-level distillation, and further improves robustness with an actionness-guided anchor decoder and anchor-level point supervision. Experiments on five benchmarks show consistent gains over strong baselines, establishing OnPoint as a competitive and label-efficient solution for online temporal action localization and a solid starting point for future POTAL research.

\section*{Acknowledgments}
This work was supported in part by the U.S. National Science Foundation (NSF) under Grant No. FW-HTF-2128743 and by the Office of Naval Research (ONR) under Grant No. N00014-21-1-2431. Any opinions, findings, conclusions, or recommendations expressed in this material are those of the authors and do not necessarily reflect the views of the NSF or ONR.
%
%
\bibliographystyle{splncs04}
\bibliography{main}

\section*{Supplementary Materials}
\appendix
\renewcommand{\thetable}{\Alph{table}}
\renewcommand{\thefigure}{\Alph{figure}}
\setcounter{figure}{0}
\setcounter{table}{0}

This supplement provides additional methodological details, analyses, and experimental results that support the main paper. 
Section~\ref{app:method} presents extended descriptions of the proposed framework, including the offline teacher setup, the distillation-free baseline, anchor-level point prediction, and the anchor decoder. 
Section~\ref{supp:exp} reports additional experimental results, including evaluations on additional datasets, post-processing comparisons, efficiency analysis, hyperparameter studies, CASS loss ablation, qualitative analysis of anchor-level point supervision, comparison of anticipation distillation strategies, evaluation using an additional metric, and attention calibration analysis.
Section~\ref{supp:annotation} investigates properties of point annotations and examines the capability of multimodal foundation models to generate pseudo temporal labels. 
Finally, we include a novelty statement (Section \ref{supp:novel}), reproducibility discussion (Section \ref{supp:repro}), and detailed implementation information (Section \ref{supp:add}) to facilitate understanding and replication of our approach.

\section{Additional Method Details}
\label{app:method}
\subsection{Offline Teacher}
Though we experimented with several offline TAL models as teacher networks in our framework (see main Table 5), we selected the best-performing one for each dataset. For the THUMOS'14 dataset, we adopted HR-Pro \cite{zhang2024hr} as the offline backbone. Rather than using its final action predictions, we extracted intermediate reliable proposals to construct pseudo ground-truth action segments for supervising our online model. In addition, the class activation sequences (CAS) produced by HR-Pro were used to supervise both the Class Activation Subsequence (CASS) predictor and the anticipation head. For the EGTEA and HOI4D-O datasets, we employed TSASPC \cite{du2023timestamp} as the offline backbone, as it is more suitable for dense procedural action scenarios where HR-Pro performs less reliably (see the offline + point comparison in main Table 2 between HR-Pro and TSASPC). The per-frame class activation outputs from TSASPC were used as CAS supervision for both the CASS predictor and the anticipation head. To generate pseudo ground-truth action instances, we grouped temporally adjacent frames belonging to the same action class into action segments. We used the original open-source implementations of HR-Pro and TSASPC for this process; additional implementation details are provided in the Implementation Details section.

\subsection{Distillation-Free Baseline}

In the distillation-free baseline, we remove anchor-level instance supervision entirely and rely solely on anchor-level point supervision as the primary training signal. During inference, we use the anchor-level point prediction for each sliding window. If the classifier activates any action class for an anchor (i.e., the activation exceeds the threshold of 0.1), we treat that anchor as a proposal. The anchor’s predefined start and end boundaries serve as the initial segment, which is then refined using the regressed center offset. The resulting adjusted segment becomes the proposal, after which we apply the same ONMS post-processing used in the full model to obtain the final detections.

A key limitation of this baseline is the lack of any scaling capability. Without pseudo ground truth, it is not possible to estimate the appropriate temporal span of a positive anchor. Consequently, the offset adjustment relies only on the heuristic observation that point annotations typically fall near the center of the action, making it an imprecise approximation. This highlights why a distillation-free approach is fundamentally limited in our point-supervised setup.

\subsection{Anchor-Level Point Prediction}

This module consists of two heads, one for classification and one for regression, each implemented as a lightweight two-layer neural network with a ReLU activation in between. Both heads take as input the anchor embedding vector of dimension $D$. The classification head outputs a probability distribution over $C$ action classes, while the regression head outputs a scalar representing the normalized temporal distance between the anchor center and the annotated point. Specifically, both heads use a first linear layer that maps from dimension $D$ to $D$, followed by a ReLU activation, and a second linear layer that maps from dimension $D$ to either $C$ (for classification) or 1 (for regression).

The classification head predicts whether a given anchor contains a point annotation and assigns the corresponding action class. The loss for this head is computed using the standard cross-entropy between the predicted class probabilities and the ground-truth class label for each annotated point. This encourages the model to assign high confidence to the correct class if the anchor is aligned with an action point.

The regression head estimates the normalized offset between the anchor center and the temporal location of the annotated point, relative to the anchor's length. The loss for this head is defined as the average absolute error between the predicted distance and the ground-truth normalized distance. 


\subsection{Anchor Decoder}

The following outlines the step-by-step implementation of our proposed actionness-guided anchor decoder.

\noindent \underline{\textbf{Anchor Decoder with Actionness-Calibrated Attention}}

\begin{enumerate}
\item \textbf{Input:} \\
Window features $F_t \in \mathbb{R}^{W \times D}$ \\
CASS prediction $A_t \in \mathbb{R}^{W \times (C+1)}$ \\
Number of anchor queries $M$, feature dimension $D$, number of decoder layers $L$
\item \textbf{Output:} \\
Anchor features $Z \in \mathbb{R}^{M \times D}$

\item \textbf{Initialize:} \\
Learnable anchor queries $\mathcal{Q} \in \mathbb{R}^{M \times D}$

\item \textbf{Compute actionness scores:}
\begin{align*}
r &= 1 - \sigma\left(A_t[:, C+1]\right) \\
\bar{r} &= r + \log(r)
\end{align*}

\item \textbf{Project window features:}
\begin{align*}
\mathcal{K} &= \text{Linear}(F_t) \in \mathbb{R}^{W \times D} \\
\mathcal{V} &= \text{Linear}(F_t) \in \mathbb{R}^{W \times D}
\end{align*}

\item \textbf{For each layer $l = 1$ to $L$:}
\begin{enumerate}
    \item Apply self-attention on anchor queries:
    \[
    \mathcal{Q} \leftarrow \texttt{SelfAttention}(\mathcal{Q})
    \]
    
    \item Compute actionness-calibrated cross-attention:
    \begin{align*}
    \text{logits} &\leftarrow \frac{\mathcal{Q} \mathcal{K}^\top}{\sqrt{D}} + \bar{r}^\top \\
    \alpha &\leftarrow \text{Softmax}(\text{logits}) \\
    \text{context} &\leftarrow \alpha \cdot \mathcal{V}
    \end{align*}

    \item Update anchor queries via residual and feedforward:
    \[
    \mathcal{Q} \leftarrow \texttt{FeedForward}(\mathcal{Q} + \text{context})
    \]
\end{enumerate}

\item \textbf{Return:} \\
$Z \leftarrow \mathcal{Q}$

\end{enumerate}

\section{Additional Experiments \& Discussion}
\label{supp:exp}

\subsection{Evaluation on Additional Datasets}
\label{sup_sec:sota}
In addition to THUMOS, EGTEA, and HOI4D reported in the main paper, we evaluate OnPoint on two additional benchmarks, EPIC-Kitchens-100 \cite{damen2022rescaling} and FineAction \cite{liu2022fineaction}, to further assess its generalization ability. These datasets contain more diverse, fine-grained, and temporally overlapping action instances, posing additional challenges for online action understanding. As shown in Table~\ref{tab:supp_sota}, OnPoint consistently outperforms existing baselines on both datasets, demonstrating the effectiveness of our approach across a broader range of action domains and complexities.

\begin{table}[h]
\small
\centering
\caption{Comparisons of performance on EPIC-Kitchen-100 and FineAction datasets. (* indicates our produced baselines.) }
\begin{tabular}{c | c|ccccc|c}
\hline
\multirow{2}{*}{Datasets} & \multirow{2}{*}{Methods} & \multicolumn{5}{c|}{mAP@tIoU (\%)} & AVG \\
& & 0.1 & 0.2 & 0.3 & 0.4 & 0.5 & [0.1:0.5] \\
\hline
EK-100 & TSASPC+OAT-ONMS* & 10.4 & 9.6 & 8.7 & 7.6 & 6.1 & 8.5 \\
 & \textbf{OnPoint (Ours)} & \textbf{12.5} & \textbf{11.8} & \textbf{10.9} & \textbf{9.7} & \textbf{7.9} & \textbf{10.5} \\
 \hline
FineAction & TSASPC+OAT-ONMS* & 7.8 & 6.7 & 5.3 & 3.9 & 2.9 & 5.3 \\
 & \textbf{OnPoint (Ours)} & \textbf{10.9} & \textbf{9.3} & \textbf{7.3} & \textbf{5.4} & \textbf{4.1} & \textbf{7.4} \\

\hline
\end{tabular}
\label{tab:supp_sota}
\end{table}

\subsection{Online Post-Processing Analysis}
\label{supp:post}
Tables \ref{tab:ablation_sup1} and \ref{tab:ablation_sup2} present a comparison between the two most commonly used online post-processing methods, OSN \cite{kim2022sliding} and ONMS \cite{song2024online}, in terms of performance and promptness. The results indicate that ONMS significantly outperforms OSN in detection accuracy but is less prompt in generating predictions. Conversely, OSN offers faster response times but at the cost of substantially lower performance. In our framework, we adopt ONMS, accepting a slight reduction in promptness for better accuracy. However, given the plug-and-play nature of our post-processing module, OSN can be used instead if promptness is a higher priority in specific use cases.






\begin{table}[h]
\scriptsize
\centering
\caption{\textbf{Performance comparison of different post-processing methods applied to our online model}. OSN refers to the Online Suppression Network proposed by \cite{kim2022sliding}, ONMS denotes the Online Non-Maximum Suppression introduced in \cite{song2024online}, and NMS is the standard offline Non-Maximum Suppression included for reference purposes only. }
\begin{tabular}{c|ccccc|c}
\hline
\multirow{2}{*}{Post-Processing} & \multicolumn{5}{c|}{mAP@tIoU (\%)} & AVG \\
& 0.1 & 0.2 & 0.3 & 0.4 & 0.5 & [0.1:0.5] \\
\hline
OSN & 63.6 & 60.0 & 55.0 & 45.9 & 33.7 & 51.6 \\

\textbf{ONMS} & \textbf{73.6} & \textbf{70.3} & \textbf{63.9} & \textbf{56.3} & \textbf{45.2} & \textbf{61.9} \\

\hline
NMS & 75.0 & 71.5 & 65.8 & 58.0 & 45.6 & 63.2 \\

\hline
\end{tabular}
\label{tab:ablation_sup1}
\end{table}

\begin{table}[h]
\scriptsize
\centering
\caption{\textbf{Promptness analysis of our framework with different online post-processing methods.} OSN refers to the Online Suppression Network proposed by \cite{kim2022sliding} and ONMS denotes the Online Non-Maximum Suppression introduced in \cite{song2024online}. Average Early Detection Time (AEDT) is the metric to measure promptness of the model proposed by \cite{kim2022sliding}. }
\begin{tabular}{c|ccccc|c}
\hline
\multirow{2}{*}{Post-Processing} & \multicolumn{5}{c|}{AEDT@IOU (sec.)} & AVG \\
& 0.1 & 0.2 & 0.3 & 0.4 & 0.5 & [0.1:0.5] \\
\hline
OSN & -1.59 & -1.57 & -1.56 & -1.52 & -1.43 & -1.53 \\

\textbf{ONMS} & -0.18 & -0.18 & -0.17 & -0.18 & -0.15 & -0.17 \\

\hline
\end{tabular}
\label{tab:ablation_sup2}
\end{table}

 \subsection{Inference Efficiency}
 \label{supp:inf}
During inference, our framework deploys only the online student model, while the offline teacher is discarded. We therefore evaluate the inference efficiency of our online model in comparison with prior online approaches. As shown in Table \ref{tab:efficiency}, OAT\cite{kim2022sliding} remains the most efficient overall, and our method achieves comparable efficiency, outperforming other state-of-the-art online models such as HAT \cite{reza2024hat} and MATR\cite{song2024online} across model size, computational complexity, and throughput.

\begin{table}[h]
\scriptsize
\centering
\caption{\textbf{Efficiency analysis of online model inference} on THUMOS. Param. reports the total parameter count (model size), GFLOPs (Giga Floating-Point Operations) quantifies hardware-independent computational complexity, and FPS measures real-time inference performance in frames per second.}
\begin{tabular}{l|ccc}
\hline
Model & Param. $\downarrow$ & GFLOPs $\downarrow$ & FPS $\uparrow$\\
\hline
 OAT \cite{kim2022sliding} & 92M & 2.75 & 355 \\
 HAT \cite{reza2024hat} & 248M & 7.09 & 161 \\
 MATR \cite{song2024online} & 191M & 7.49 & 206 \\
 \textbf{Ours} & 93M & 2.88 & 312 \\
\hline
\end{tabular}

\label{tab:efficiency}
\end{table}

Here, the cost of the pre-trained feature extractor is excluded, as it is identical across all methods. Inference speed is measured on an NVIDIA RTX 4090 GPU.

\subsection{Hyperparameter Analysis}

\subsubsection{Loss-Weight Hyperparameter Analysis.}The overall training objective is defined as a weighted combination of four loss terms,
$\mathcal{L}_{\text{total}} = \alpha \mathcal{L}_{\text{ins}} + \beta \mathcal{L}_{\text{cass}} + \gamma \mathcal{L}_{\text{ant}} + \delta \mathcal{L}_{\text{pnt}}$,
where $\mathcal{L}_{\text{ins}}$ is the anchor-level instance prediction loss, $\mathcal{L}_{\text{cass}}$ the class-activation sub-sequence loss, $\mathcal{L}_{\text{ant}}$ the anticipation loss, and $\mathcal{L}_{\text{pnt}}$ the point prediction loss. Detailed definitions of these components are provided in the main paper.

Table \ref{tab:hyp} presents the sensitivity of our method to the loss-weight hyperparameters. On THUMOS, we find that assigning equal weights $\alpha=\beta=\gamma=\delta=1$ yields the strongest performance. Reducing any individual weight leads to a consistent drop in accuracy. For the other two datasets, we observe a slight improvement when setting $\gamma = 0.8$, but the gain is marginal.

Overall, these results show that deviating from the default hyperparameter settings provides little to no benefit. Thus, we could adopt the simple and robust configuration of setting all loss weights to 1, which eliminates unnecessary tuning effort while preserving strong and consistent performance across all datasets.

\begin{table}[h]
\scriptsize
\centering
\caption{\textbf{Hyperparameter analysis} conducted on the THUMOS dataset.}
\begin{tabular}{cccc|c}
\hline
 \multicolumn{4}{c|}{Hyperparameters} & AVG mAP \\
 $\alpha$ & $\beta$ & $\gamma$ & $\delta$ & tIoU@[0.1:0.5] \\
\hline
 \textbf{1.0} & \textbf{1.0} & \textbf{1.0} & \textbf{1.0} & \textbf{61.9} \\
 \hline
 \textbf{0.8} & 1.0 & 1.0 & 1.0 & 60.2 \\
 \textbf{0.6} & 1.0 & 1.0 & 1.0 & 59.2 \\
 \textbf{0.4} & 1.0 & 1.0 & 1.0 & 59.1 \\
 \hline
 1.0 & \textbf{0.8} & 1.0 & 1.0 & 60.0 \\
 1.0 & \textbf{0.6} & 1.0 & 1.0 & 58.5 \\
 1.0 & \textbf{0.4} & 1.0 & 1.0 & 58.2 \\
\hline
 1.0 & 1.0 & \textbf{0.8} & 1.0 & 60.7 \\
 1.0 & 1.0 & \textbf{0.6} & 1.0 & 60.5 \\
 1.0 & 1.0 & \textbf{0.4} & 1.0 & 59.0 \\
\hline
 1.0 & 1.0 & 1.0 & \textbf{0.8} & 59.3 \\
 1.0 & 1.0 & 1.0 & \textbf{0.6} & 59.6 \\
 1.0 & 1.0 & 1.0 & \textbf{0.4} & 59.4 \\
\hline
\end{tabular}
\label{tab:hyp}
\end{table}

\subsubsection{Anchor Hyperparameter Analysis.} 
We study the effect of the number and lengths of temporal anchors. 
Following OAT~\cite{kim2022sliding}, we set the window size and maximum anchor length to 64, which covers approximately 96\% of action instances in THUMOS. 
Table~\ref{tab:anchor_ablation} reports the results for different configurations. 
The best performance is obtained with $M=6$ anchors $\{4,8,16,32,48,64\}$, achieving $61.9\%$ average mAP. 
Using fewer anchors or smaller maximum anchor length degrades performance, while increasing the number of anchors or the maximum anchor length does not provide further improvement, indicating a clear optimum with moderate sensitivity to anchor design.

\begin{table}[h]
\centering
\small
\caption{Effect of the number and lengths of temporal anchors on THUMOS. }
\begin{tabular}{c c c}
\hline
\# Anchors ($M$) & Anchor Lengths & Avg. mAP (\%) \\
&& tIoU@[0.1:0.5] \\
\hline
4 & $\{4,8,16,32\}$ & 58.7 \\
4 & $\{8,16,32,64\}$ & 59.2 \\
\textbf{6} & $\mathbf{\{4,8,16,32,48,64\}}$ & \textbf{61.9} \\
8 & $\{4,8,16,24,32,64,96,128\}$ & 59.8 \\
8 & $\{4,8,12,16,24,32,48,64\}$ & 60.0 \\
\hline
\end{tabular}
\label{tab:anchor_ablation}
\end{table}

\subsection{CASS Loss}
Table~\ref{tab:cass_loss} compares different loss functions for CASS alignment. Since CASS provides multi-label, per-time-step supervision with independent per-frame actionness scores, losses that require normalization (e.g., distribution-based losses such as KL divergence) are not well suited. Regression-style losses such as L1 (59.8\%) and Huber (59.0\%) preserve the multi-label semantics but are less effective at aligning peak activation magnitudes. Cosine similarity (60.5\%) improves temporal shape alignment but ignores the absolute strength of activations. BCE (61.7\%) performs strongly by modeling independent Bernoulli supervision for each class–time pair. Although distribution-based losses are not principled for this setting, we also include KL divergence (60.0\%) by forcibly normalizing the activations for experimental completeness; this inappropriate normalization leads to degraded performance. Overall, L2 (MSE) achieves the best performance (61.9\%), suggesting that directly aligning continuous actionness magnitudes without normalization provides the most stable and effective supervision.

\begin{table}[h]
\centering
\small
\caption{Comparison of different loss functions for CASS alignment on THUMOS.(*with forced normalization)}
\begin{tabular}{l | c}
\hline
Loss Function & Avg mAP (\%) \\
& tIoU@[0.1:0.5] \\
\hline
L1 Loss & 59.8 \\
\textbf{L2 Loss} & \textbf{61.9} \\
Huber Loss & 59.0 \\
KL Divergence* & 60.0 \\
Cosine Similarity & 60.5 \\
BCE Loss & 61.7 \\
\hline
\end{tabular}

\label{tab:cass_loss}
\end{table}



\subsection{Anchor-Level Point Supervision - Qualitative Analysis} 
\begin{figure}[h]
	\centerline{\includegraphics[width=0.9\linewidth]{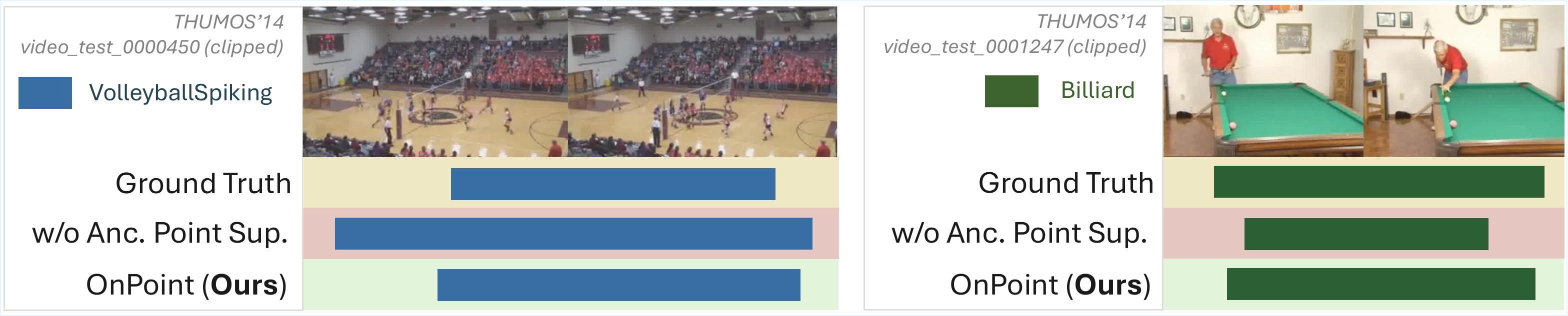}}
    \vspace{-2mm}
    \caption{Qualitative analysis: Anchor-level point supervision} \label{fig:rebut_qual}
\end{figure}

Qualitative results on THUMOS (Fig. \ref{fig:rebut_qual}) demonstrate that anchor-level point supervision in the student yields more accurate instance-level localization.

\subsection{Comparison of Window-Level Anticipation Distillation Strategies}
\label{supp:anti}

The anticipation window size $W'$ is critical for effective future distillation. A small $W'$ provides limited anticipation and largely overlaps with the ongoing action, while a large $W'$ introduces distant future events that dilute the supervision signal. As demonstrated in Fig.~\ref{main:anti} of the main paper on THUMOS, performance peaks at $W'=16$ (61.9\% Avg mAP) and gradually declines for both smaller and larger window sizes. We further compare this tuned fixed-window strategy with several alternatives, including a multi-scale window ($W' \in {8,16,24}$), a CASS-conditioned adaptive window, and a next-transition prediction objective. As shown in Table~\ref{tab:supp_anti}, these variants achieve 60.8\%, 60.4\%, and 61.6\% Avg mAP, respectively, all matching or underperforming the fixed-window strategy, indicating that a properly chosen fixed anticipation horizon provides the most effective supervision for window-level anticipation distillation.

\begin{table}[h]
\centering
\small
\caption{Comparison of different anticipation strategies for action anticipatory distillation.}
\begin{tabular}{l | c}
\hline
Methods & Avg mAP (\%) \\
& tIoU@[0.1:0.5] \\
\hline
 Fixed $W'= 16$ & \textbf{61.9}  \\
 Multi-Scale $W'= \{8,16,24\}$ & 60.8  \\
 CASS-Conditioned Adaptive $W'$ & 60.4  \\
 Next-Transition Prediction & 61.6  \\
\hline
\end{tabular}

\label{tab:supp_anti}
\end{table}

\subsection{Additional Evaluation Metric}

In addition to the standard mAP@tIoU metric reported in the main paper, we further evaluate our method using the instance-level F1 score at tIoU = 0.5, following \cite{kang2024actionswitch}. As shown in Table~\ref{tab:supp_f1}, OnPoint consistently outperforms the baseline methods across different datasets under this complementary evaluation metric, demonstrating its ability to produce more accurate and reliable action instance predictions.

\begin{table}[h]
\scriptsize
\centering
\caption{Comparison of instance-level F1 performance across datasets. (* indicates our produced baselines.) }
\begin{tabular}{c|c|c}
\hline
Dataset & Method & F1 (\%) \\

\hline
THUMOS'14 & HR-Pro + OAT-ONMS* & 51.3 \\

& \textbf{OnPoint (Ours)} & \textbf{55.6} \\

\hline
HOI4D-O & TSASPC + OAT-ONMS* &  41.7\\

& \textbf{OnPoint (Ours)} & \textbf{44.7} \\

\hline
EK-100 & TSASPC + OAT-ONMS* & 11.1 \\

& \textbf{OnPoint (Ours)} & \textbf{13.9} \\

\hline

FineAction & TSASPC + OAT-ONMS* & 7.6 \\

& \textbf{OnPoint (Ours)} & \textbf{10.1} \\

\hline
\end{tabular}
\label{tab:supp_f1}
\end{table}

\subsection{Attention Calibration Analysis}
\label{supp:calib}

To better understand the effect of attention calibration, we analyze the modified attention score $\bar{r}_t=\log(r_t)+r_t$, where $r_t\in[0,1]$ denotes the estimated action relevance of the $t$-th key token. Since $\bar{r}_t$ is added to the pre-softmax attention logits, it biases attention toward action-relevant tokens by suppressing low-relevance ones while emphasizing highly relevant ones. To quantify the resulting attention behavior, we analyze attention distributions on THUMOS. The calibrated model increasingly concentrates attention on action regions, yielding an overall upward trend in the action-to-background attention-mass ratio across layers (slope $+0.075$/layer versus $-0.015$/layer for the baseline) and reaching $1.11$ at the final layer compared to $0.93$, corresponding to an approximately $19\%$ relative improvement. Furthermore, it maintains lower attention entropy throughout the network (mean $2.83$ versus $3.23$ nats), indicating sharper and more selective attention patterns, consistent with the qualitative visualization in the main paper (Fig.~\ref{fig:attention_calib}). Together, these results suggest that attention calibration improves action-background discriminability and produces more action-focused representations that contribute to stronger localization performance.

\section{Annotation Analysis}
\label{supp:annotation}

\subsection{Point Annotation Analysis}
To understand how point annotations are collected by human annotators in realistic settings, how much annotation time they save, and what their distribution looks like, we combine insights from prior work \cite{ma2020sf} with findings from our own annotation collection. Specifically, we conducted a small-scale annotation study with 14 annotators. We selected 10 THUMOS video clips and divided them into two sets with equal numbers of videos and roughly equal total duration. For each annotator, one set was randomly assigned for point annotations, and the other set was used for full temporal segment annotations.

\subsubsection{Point-Annotation Distribution.} 
\begin{figure}[t]
\centering
\includegraphics[width=0.7\columnwidth]{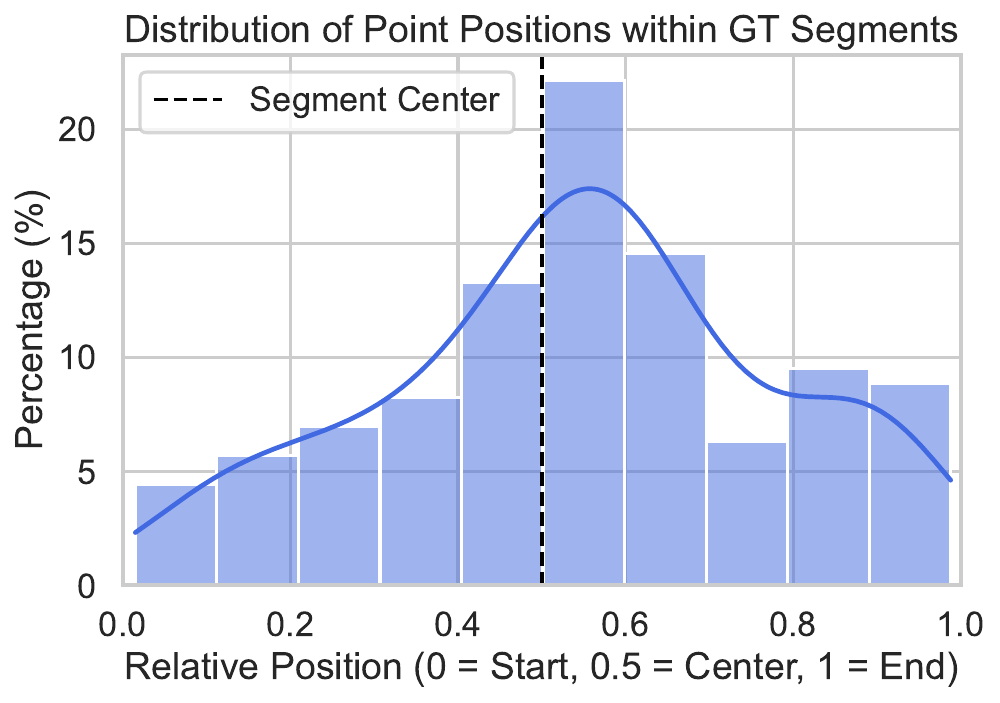} 
\caption{Point annotation distribution relative to normalized action instance length (0–1 scale), aggregated from 14 annotators on sample THUMOS videos.}
\label{fig:point}
\vspace{-0.5cm}
\end{figure}

Consistent with observations in \cite{ma2020sf}, our collected annotations also cluster near the center of the action instances, forming a distribution that closely resembles a Gaussian over normalized temporal position. As illustrated in Fig. \ref{fig:point}, annotators naturally gravitate toward mid-action frames when providing a single temporal point, suggesting that point labels convey semantically meaningful cues aligned with the core of the action.

\subsubsection{Point-Annotation Time Savings.} The prior work \cite{ma2020sf} quantified annotation cost across different supervision levels using the GTEA dataset. Four trained annotators labeled 1-minute videos and required 50 seconds per clip for point annotations compared to 300 seconds for full temporal segments, yielding a 6× speed-up. In our annotation data collection on THUMOS, annotators required 1.6× less time for point annotations than for full annotations. Together, these results highlight the substantial reduction in annotation effort offered by point supervision, while still enabling performance close to (and in some cases exceeding) fully supervised methods.

\subsection{Multimodal Foundation Model Labeling Analysis.} 
To evaluate the potential of large multimodal foundation models for automatic video annotation in temporal localization tasks, we use Gemini 2.5 Flash \cite{comanici2025gemini} as a representative high-end multimodal model. We prompt the model to generate pseudo labels for two types of supervision: (1) full temporal segments and (2) point-level annotations. The prompts used for these two settings are provided in Table \ref{tab:gemini_prompts}.

We evaluate the generated annotations on the THUMOS14 test set. For full temporal segments, Gemini achieves 29.2\% average mAP@[0.1:0.5], compared to 90.3\% obtained using pseudo full-segment labels generated by our offline teacher model. For point supervision, Gemini produces pseudo point labels with 30.9\% mAP, while in our study we found that human point annotation achieves 81.6\% mAP.

\begin{figure}[!h]
\centering

\includegraphics[width=\columnwidth]{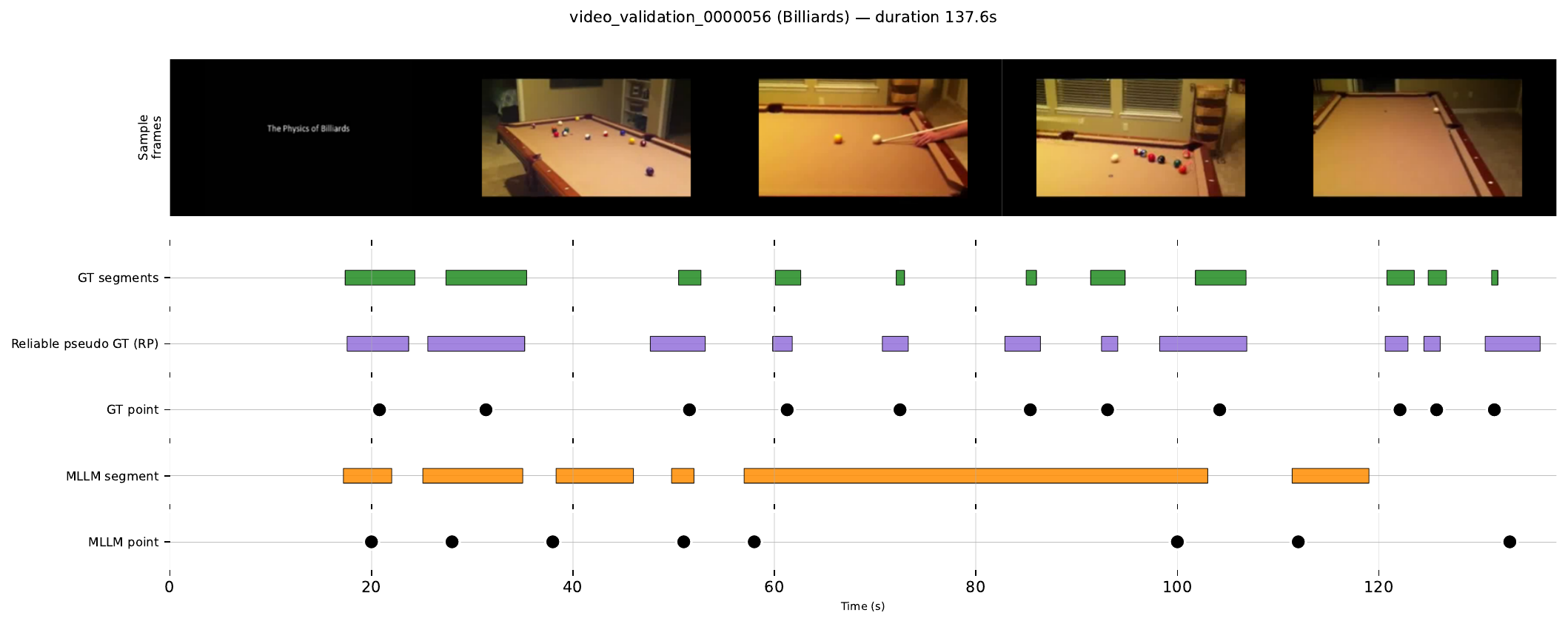}

\vspace{0.2cm}

\includegraphics[width=\columnwidth]{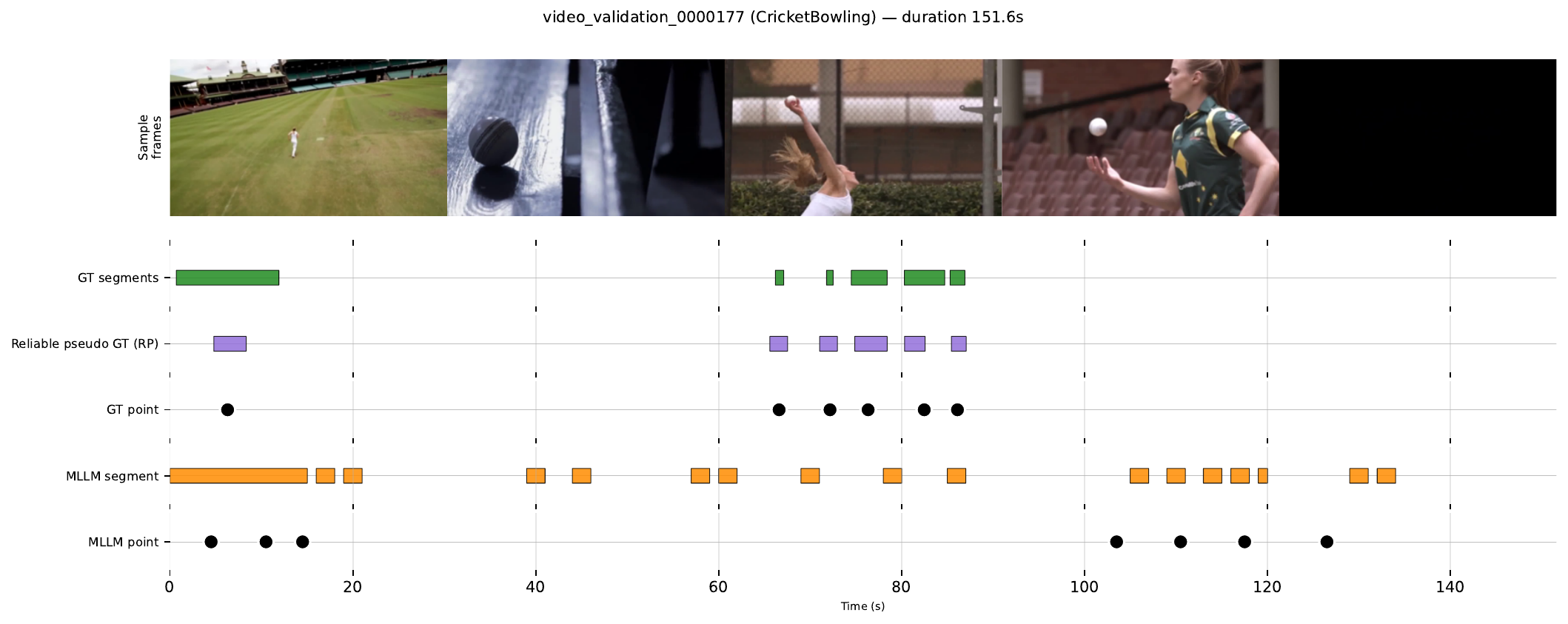}

\vspace{0.2cm}

\includegraphics[width=\columnwidth]{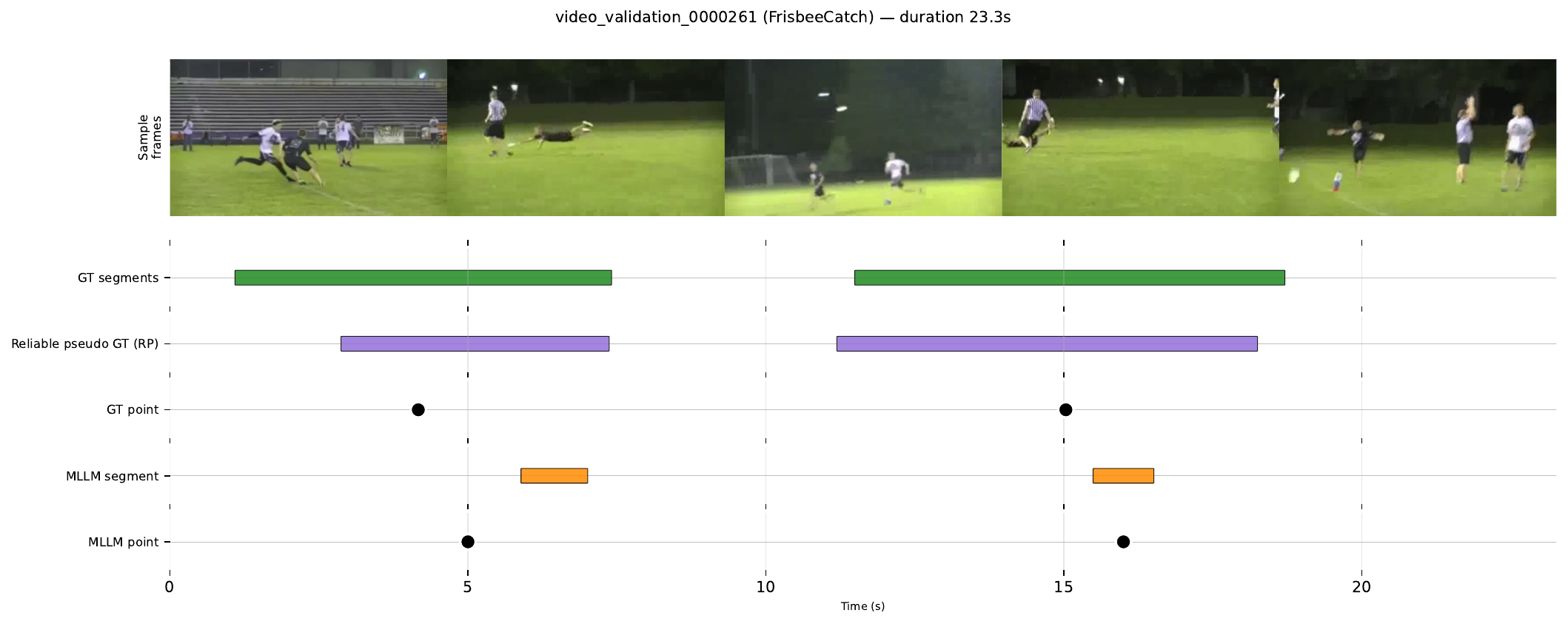}

\caption{Qualitative examples on THUMOS'14 comparing annotations generated by the multimodal LLM (Gemini 2.5 Flash) for both full segments and point annotations with pseudo ground truth produced by our offline teacher, along with the corresponding ground truth point and full segment annotations.}
\label{fig:gemini_examples}
\vspace{-0.5cm}

\end{figure}

These results indicate that off-the-shelf multimodal foundation models are still not sufficiently reliable for generating pseudo supervision for temporal action localization, particularly when compared with task-specific models trained using weak human labels. Additional qualitative examples are shown in Figure~\ref{fig:gemini_examples}.

\section{Novelty Statement}
\label{supp:novel}

\paragraph{Point-Supervised Online Temporal Action Localization (POTAL).}
While prior work has explored point-supervised temporal action localization (PS-TAL) in offline settings and Online Temporal Action Localization (OnTAL) separately, no existing effort has attempted to bridge the label-efficiency of point annotations with the online streaming constraints of OnTAL. We introduce POTAL, the first formulation of point-supervised online temporal action localization, where we define the problem, establish a benchmark, and develop both distillation-based and distillation-free baselines tailored for this new setting.


\paragraph{Offline-to-Online Multi-Level Distillation.}
Although distillation has been explored across different tasks and modalities,
including some forms of offline-to-online distillation in other domains such as
instance segmentation \cite{kim2024offline}, spatio-temporal action detection
\cite{patel2026distilling}, language modeling \cite{liu2025offline}, and
reinforcement learning \cite{li2024guided}, no prior work distills knowledge from
an offline TAL model into an online temporal action localization model. The
closest of these, DSTA \cite{patel2026distilling}, distills an offline
\emph{action detection} model into a streaming per-frame detector by aligning
spatial attention and RoI features between architecturally similar teacher and
student detectors; in contrast, OnPoint distills temporal supervision (pseudo
segments, frame-level CASS, and window-level anticipation) from an
architecturally distinct offline TAL teacher that produces full action
\emph{segments} rather than per-frame labels. Our framework introduces a novel
multi-level distillation design composed of three complementary components.
Among them, anchor-level instance distillation using pseudo ground truth is an
intuitive choice. However, the incorporation of Class-Activation Sub-Sequence
(CASS) distillation, and furthermore, using the offline CAS to guide anchor
decoding within each online window, is a new and non-obvious idea. In addition,
we propose window-level anticipation distillation, a tailored mechanism that
enables the online model to anticipate upcoming action evolution based on offline
temporal context. Together, these components form a unique offline-to-online
distillation framework purpose-built for the POTAL setting.

\paragraph{Class-Activation Sub-Sequence Distillation.}

Class-Activation Sequences (CAS) have been widely used in offline weakly supervised TAL for producing proposals, but they have never been integrated into an online distillation pipeline.
We propose CASS distillation, where the online model’s sliding-window CAS is aligned to the corresponding subsequence of the offline CAS. This provides fine-grained, frame-level supervision within each window, representing the first adoption of CAS-based distillation for online temporal modeling.

\paragraph{Window-Level Action Anticipatory Distillation.} 
Although anticipation has appeared as a primary or auxiliary objective in several action understanding tasks \cite{guermal2024joadaa, wang2023memory}, 
no prior work has leveraged anticipation within a distillation framework for online action localization.
Our window-level anticipatory distillation explicitly teaches the online model to predict upcoming action dynamics, enabling more accurate and responsive boundary localization in streaming video.

\paragraph{Auxiliary Anchor-Level Point Annotation.}
Point annotation has been applied in offline TAL and other areas such as object detection \cite{chen2021points}.
However, this is the first work to design a point-based supervisory signal directly used for an online action localization model. We introduce a new anchor-level point supervision objective, tailored to the anchor-based decoding structure of POTAL, which differs fundamentally from previous point-annotation formulations.

\paragraph{Actionness-Guided Anchor Decoder.} 
Actionness scores have appeared in the literature for temporal localization and action proposal generation. However, to our knowledge, no prior work uses actionness to calibrate attention inside a transformer decoder for anchor feature construction. We propose an actionness-guided anchor decoder, where actionness modulates the cross-attention weights to emphasize informative temporal regions, improving anchor-level feature quality in the online setting.

\section{Reproducibility Statement}
\label{supp:repro}
Due to an active Non-Disclosure Agreement (NDA), we are currently unable to release our full source code. However, we have taken comprehensive steps to ensure that our work is fully reproducible. We provide detailed descriptions of our methodology, complete implementation details, and links to all open-source components used to build our framework. For our novel components without publicly available code, we include step-by-step descriptions or pseudocode to facilitate reproduction. Additionally, we document all hyperparameters, as well as our hardware and software environment configurations, to support transparent and consistent reproduction of our experiments.

\section{Additional Implementation Details}
\label{supp:add}
\subsection{Hardware System Configuration}
For all experiments, we used a server equipped with an NVIDIA L4 GPU running NVIDIA Driver Version 535.183.06 and 24 GB of memory. For the inference efficiency analysis only, we used a workstation with an NVIDIA RTX 4090 GPU, as the server was temporarily unavailable; however, the evaluation was performed using the same trained model checkpoint from the original setup.

\subsection{Software Environment Configuration}
Our experiments were conducted using a Conda environment. The environment included core dependencies from the \texttt{pytorch}, \texttt{nvidia}, and \texttt{defaults} channels. Key packages included Python 3.10.18, PyTorch 2.5.1 with CUDA 11.8, TorchVision 0.20.1, Torchaudio 2.5.1, and supporting CUDA libraries (\texttt{libcublas}, \texttt{libcusparse}, \texttt{libnvjpeg}, etc.). MKL-based NumPy (2.0.1), SciPy (1.15.3), and Pandas (2.3.1) were used for numerical and data processing tasks. Visualization and logging were supported by Matplotlib (3.10.3), TensorBoard (2.19.0), TensorBoardX (2.6.4), and Rich (14.0.0). For machine learning and deep learning, additional packages included TensorFlow 2.19.0, Keras 3.10.0, Scikit-learn dependencies like Joblib (1.5.1), and optimization utilities such as Opt-einsum (3.4.0). The environment also included standard libraries such as \texttt{cffi}, \texttt{openssl}, \texttt{sqlite}, and compression tools (\texttt{zlib}, \texttt{xz}, \texttt{lz4}). All packages were installed either through Conda or via \texttt{pip}, ensuring reproducibility across experiments. We used fixed random seeds across all experimental runs to ensure reproducibility. Specifically, the online model was run with a seed of 52. For the offline model based on HR-Pro (used on THUMOS'14), we used a seed of 0, and for the TSASPC-based offline model (used on EGTEA and HOI4D-O), we used a seed of 1538574472. Consistent with prior work, all experiments were conducted as single runs using these fixed seeds.

\subsection{Generating Ground Truth for Online Model}
For the THUMOS'14 dataset, we used HR-Pro as the offline backbone. Rather than using the final output action predictions, we extracted the intermediate reliable proposals to serve as pseudo ground truth for training the online model. Additionally, we used the class activation sequences (CAS) from HR-Pro to supervise both the Class Activation Subsequence (CASS) predictor and the anticipation head. The implementation of HR-Pro is available at \textcolor{blue}{\url{https://github.com/pipixin321/HR-Pro}}. 

For the EGTEA and HOI4D-O datasets, we adopted TSASPC as the offline backbone. We utilized the per-frame class activation outputs from TSASPC as the CAS ground truth for supervising the CASS predictor and anticipation head. To generate action instance pseudo ground truth, we grouped temporally adjacent frames of the same action class. The implementation for TSASPC can be found at \textcolor{blue}{\url{https://github.com/ddz16/TSASPC/}}.

\subsection{Online Model Implementation}
The online model implementation builds upon OAT, with the original OAT code available at \textcolor{blue}{\url{https://github.com/YHKimGithub/OAT-OSN/}}. 
Additional components, such as the Anticipation Head, CASS Predictor, Actionness-Calibrated Attention in the Anchor Decoder, and Anchor-Level Point Prediction, are detailed in the main paper and earlier sections of this supplementary material, enabling full reproduction of our method. For online post-processing, we used ONMS, with its implementation available at \textcolor{blue}{\url{https://github.com/skhcjh231/MATR_codebase/}}.

\subsection{Datasets, Pre-Extracted Feature, and Labels}
For the THUMOS'14 dataset, we used the pre-extracted I3D features and point labels from the HR-Pro GitHub repository for the offline teacher model, and the pre-extracted features from the OAT GitHub repository for the online student model. For EGTEA, the same pre-extracted features were used for both the offline and online models from  HAT \cite{reza2024hat}'s GitHub Repo - \textcolor{blue}{\url{https://github.com/sakibreza/ECCV24-HAT}}. 
For HOI4D-O, we followed the feature extraction procedure described in \cite{reza2023enhancing} to generate features. For both EGTEA and HOI4D-O, point annotations were generated by sampling a single frame from a Gaussian distribution centered at the midpoint of each action instance, with a standard deviation equal to one-sixth of the instance duration.

Following prior OnTAL work \cite{kim2022sliding, reza2024hat, song2024online}, we do not include ActivityNet v1.3 \cite{caba2015activitynet}
in our evaluation. As noted in \cite{kim2022sliding}, ActivityNet v1.3 is not well-suited for the Online TAL setting because its videos typically contain only a single action instance that spans most of the video duration. This contradicts the primary objective of On-TAL, which is to detect multiple, potentially overlapping action instances in a streaming environment.

\subsubsection{Hyperparameters}
For offline teacher pre-training, we follow the exact configuration and hyperparameters provided in the HR-Pro GitHub repository for the THUMOS14 dataset. For EGTEA and HOI4D-O, we adopt the TSASPC model with the following settings: number of stages = 4, number of layers = 10, number of feature maps = 64, feature dimension = 2048, batch size = 8, and learning rate = 0.0005. For the online model across all datasets, we use a batch size of 128 during training. The model architecture includes a transformer encoder with a hidden dimension of 1024, consisting of 3 blocks with 8 attention heads each, and an anchor decoder composed of 5 transformer blocks with 4 heads per block. We apply a dataset-specific sliding window setup and anchor configuration: for THUMOS14, we use a window size of 64, 6 anchors with lengths {4, 8, 16, 32, 48, 64}, and an anticipation window of 16; for EGTEA, a window size of 24, 7 anchors with lengths {2, 4, 6, 8, 12, 16, 24}, and an anticipation window of 12; and for HOI4D-O, a window size of 56, 6 anchors with lengths {4, 8, 16, 24, 48, 56}, and an anticipation window of 16. The anticipation loss weight $\gamma$ is set to 1.0 for THUMOS14 and 0.8 for EGTEA and HOI4D-O. We train all models using the Adam optimizer with a learning rate of 1e-4 and weight decay of 1e-4.

\subsubsection{Evaluation Metrics} We follow prior work and report mean Average Precision (mAP) at multiple IoU thresholds (e.g., 0.1, 0.2, 0.3, 0.4, 0.5) to evaluate the temporal alignment and classification accuracy of predicted action segments. A prediction is considered correct if it matches the ground truth action label and the Intersection over Union (IoU) between the predicted and ground truth segments exceeds the specified threshold. For each IoU threshold, we compute the Average Precision (AP) per class and then take the mean across all classes to obtain mAP@X. This multi-threshold evaluation provides a comprehensive assessment of the model’s localization performance, from coarse to fine granularity.

\begin{table}[!h]
\centering
\small
\caption{Prompts used with Gemini 2.5 Flash for generating pseudo labels: (a) full segment annotations and (b) point annotations.}
\begin{tabular}{p{0.95\columnwidth}}
\hline
\textbf{(a) Full Segment Label Generation Prompt} \\

In the given video, there may be multiple instances of the following actions:
\textit{[Action Classes]}. Instances may overlap. Detect each action instance with the class label,
start and end time in seconds, and confidence score. Maintain the specified
output JSON format:

\begin{verbatim}
[
  {
    "label": class_label,
    "segment": [
      start_time_in_seconds,
      end_time_in_seconds
    ],
    "score": confidence_score
  }
]
\end{verbatim}
\\
\hline

\textbf{(b) Point Label Generation Prompt} \\

In the given video, there may be multiple instances of the following actions:
\textit{[Action Classes]}. Detect each action instance with the class label, a single timestamp
(in seconds) that falls approximately in the middle of the action instance's
start and end times, and confidence score. Maintain the specified output
JSON format:

\begin{verbatim}
[
  {
    "label": class_label,
    "point": timestamp_in_seconds,
    "score": confidence_score
  }
]
\end{verbatim}
\\
\hline
\end{tabular}

\label{tab:gemini_prompts}
\end{table}

\end{document}